\documentclass[10pt,journal,compsoc]{IEEEtran}
\ifCLASSOPTIONcompsoc
  \usepackage[nocompress]{cite}
\else
  \usepackage{cite}
\fi
\ifCLASSINFOpdf
  \usepackage[pdftex]{graphicx}
  \usepackage{wrapfig}
  \usepackage{subfigure}
\else
\fi
\usepackage{multirow}
\usepackage{adjustbox}
\usepackage{color}
\usepackage{amsmath}
\usepackage{amssymb}
\usepackage{algorithmic}
\usepackage{algorithm}

\usepackage{multirow}

\ifCLASSOPTIONcompsoc
  \usepackage[caption=false,font=footnotesize,labelfont=sf,textfont=sf]{subfig}
  \usepackage[justification=centering]{caption}
\else
  \usepackage[caption=false,font=footnotesize]{subfig}
\fi

\usepackage{caption} 

\begin{document}
%
\title{MPR-Net:Multi-Scale Pattern Reproduction Guided Universality Time Series Interpretable Forecasting}
%
%
%
%

\author{Tianlong Zhao, Xiang Ma, Xuemei Li, Caiming Zhang
        
\IEEEcompsocitemizethanks{\IEEEcompsocthanksitem Tianlong Zhao, Xiang Ma, Xuemei Li(corresponding author) and Caiming Zhang are with school of software, Shandong University, China.\protect\\
Caiming Zhang is also with Shandong Provincial Laboratory of Future Intelligence and Financial Engineering, China, and Digital Media Technology Key Lab of Shandong Province, China.\protect\\
	
E-mail: tianlongzhao@mail.sdu.edu.cn; xmli@sdu.edu.cn
}
\thanks{This work has been submitted to the IEEE for possible publication. Copyright may be transferred without notice, after which this version may no longer be accessible.}}

%
%

\markboth{Journal of \LaTeX\ Class Files,~Vol.~14, No.~8, August~2015}%
{Shell \MakeLowercase{\textit{et al.}}: Bare Demo of IEEEtran.cls for Computer Society Journals}
%



\IEEEtitleabstractindextext{%
\begin{abstract}
Time series forecasting has received wide interest from existing research due to its broad applications and inherent challenging. The research challenge lies in identifying effective patterns in historical series and applying them to future forecasting. Advanced models based on point-wise connected MLP and Transformer architectures have strong fitting power, but their secondary computational complexity limits practicality. Additionally, those structures inherently disrupt the temporal order, reducing the information utilization and making the forecasting process uninterpretable. To solve these problems, this paper proposes a forecasting model, MPR-Net. It first adaptively decomposes multi-scale historical series patterns using convolution operation, then constructs a pattern extension forecasting method based on the prior knowledge of pattern reproduction, and finally reconstructs future patterns into future series using deconvolution operation. By leveraging the temporal dependencies present in the time series, MPR-Net not only achieves linear time complexity, but also makes the forecasting process interpretable. By carrying out sufficient experiments on more than ten real data sets of both short and long term forecasting tasks, MPR-Net achieves the state of the art forecasting performance, as well as good generalization and robustness performance. 
\end{abstract}

\begin{IEEEkeywords}
Interpretability Forecasting, Multi-scale Pattern Processing, Pattern Matching\&Extension, Correlation Attention.
\end{IEEEkeywords}}

\maketitle

\IEEEdisplaynontitleabstractindextext

%
\IEEEpeerreviewmaketitle

\IEEEraisesectionheading{\section{Introduction}\label{sec:introduction}}

%
%
%
%
\IEEEPARstart{T}{ime} series consists of observations collected at regular intervals over time, exhibiting diverse patterns such as trends, seasonality, cycles, and irregular fluctuations. Time series forecasting is a valuable technique that involves analyzing historical values of a variable to discern patterns aiding in predicting its future behavior. The significance of time series forecasting lies in its ability to offer insights into the future behavior of variables based on their past observations. This information enables decision-makers to anticipate forthcoming patterns, identify potential risks, and make well-informed decisions. Consequently time series forecasting is widely used in many fields to forecast a wide range of variables such as climate change \cite{karevan2020transductive}, disease transmission \cite{matsubara2014funnel}, energy and electricity consumption \cite{pang2022hierarchical}, and traffic pressure\cite{ma2021short}. 

However, time series forecasting is a challenging task due to several difficulties that can affect the accuracy of the forecasts. 1.\textbf{Non-stationarity} \cite{liu2022non}: Time series data often exhibit non-stationarity, wherein the statistical properties such as mean and variance change over time. This poses a challenge as traditional statistical methods, which assume stationarity, are insufficient for modeling and forecasting such data accurately. 2.\textbf{Complexity} \cite{wu2022timesnet}: Time series data often display multiple patterns, including trend, seasonality, cycles, and irregular fluctuations, which may operate on variety time scales (e.g., days, months, years). Capturing and accurately modeling these intricate patterns becomes arduous, particularly when they interact in complex ways and are subject to various sources of uncertainty. 3.\textbf{Forecasting horizon} \cite{wu2021autoformer}: The accuracy of the forecasts decreases with an increase in forecasting horizon. Thus, longer-term forecasting exhibits significant lower precision than short-term forecasting. To overcome these difficulties, it is essential to employ suitable forecasting methods capable of comprehending the complexity of the data and accounting for non-stationarity. These methods should effectively capture the underlying patterns inherent in the time series, thereby improving forecast accuracy across with varying forecasting horizons.

Time series data often exhibit complex patterns that are difficult to capture using traditional statistical methods, while deep learning(DL) techniques have demonstrated great potential for overcoming these challenges. Existing DL based researches have focused on developing new deep learning architectures and techniques for time series forecasting in order to improve the performance of existing models. Generally, these researches can be categorized into two primary parts: the extraction of historical series patterns (Encoder) and the generation of future series forecasts based on the extracted patterns (Decoder).

Extracting historical series patterns is first a critical step in time series forecasting. It involves transforming raw data into meaningful patterns that deep learning algorithms can utilize for accurate forecasting. Time series variables exhibit diverse and complex patterns over different time scales, necessitating models capable of identifying and processing these patterns simultaneously. However, this approach can result in complex models and challenges in accurately capturing and forecasting time series patterns. To address these challenges, decomposing time series data \cite{cleveland1990stl} is utilized to separate different patterns into their underlying components, improving data understanding and modeling. Trend-period pattern decomposition \cite{wen2019robuststl} is widely used in deep learning for time series forecasting. The trend pattern represents the long-term progression of the series, while the periodic pattern captures its recurring nature, enabling forecasting of future patterns. Existing methods for decomposing trend patterns include sliding average \cite{wu2021autoformer,zhou2022fedformer}, exponential smoothing, and polynomial fitting\cite{oreshkin2019n}. However, these methods are sensitive to parameter setting independent of the data being predicted, impacting their effectiveness. For decomposing period patterns, traditional approaches such as Fourier decomposition and wavelet decomposition are commonly employed. However, these methods may not be suitable for all types of data with diverse periodic patterns, posing challenges in accurate modeling. Therefore, a key research challenge is to adopt appropriate decomposition methods that yield pattern containing trend and periodic components and suitable for different data sets.

Future series forecasting based on historical series pattern features is the another critical step. In term of the DL model forecasting module adopted, Recurrent Neural Networks (RNNs) are a class of DL methods that are designed to handle sequential data by allowing information to transmit over time. However, RNNs face challenges such as vanishing gradients \cite{pascanu2013difficulty} and difficulty in parallel modeling long-range dependencies. Although variants like Long Short-Term Memory (LSTM) can partially alleviate these issues, they do not provide a fundamental solution. In recent works, the point-wise connection structure has gained popularity as a forecasting module, demonstrated by models like attention-based Transformer method \cite{vaswani2017attention} and fully connected network-based multilayer perceptron method (MLP). 
Unlike RNNs, the point-wise connection structure processes input sequences in parallel, and offers strong fitting and expressive power. It reduces the theoretical path length to $O(1)$ between every sequence points, effectively modeling long-term dependencies and addressing the gradient vanishing problem associated with RNNs. However, when time series has a large number of time steps, this structure can result in high-dimensional input spaces, increasing the number of parameters and resulting in quadratic computational complexity. This can lead to overfitting, particularly with limited available data, and may also suffer from the curse of dimensionality, hindering effective learning and generalization. 
Moreover, while the point-wise connection structure reduces the theoretical path length between sequence points, it discards the temporal dependencies present in the time series \cite{zhao2023asset} and processes each input data point independently. The loss of sequential temporal relationship between points potentially reduces the information utilization and forecast accuracy. Although location and temporal encoding \cite{li2019enhancing} can alleviate this issue, the network's fundamental structure limitations prevent a complete resolution.
Another problem with ignoring temporal dependencies between points is the lack of interpretability of the forecasting process. Interpretability plays a vital role in understanding the forecasting process \cite{rudin2019stop} based on historical pattern features, which is crucial for accurate forecasting. The lack of interpretability hampers trust in the model forecasting, making it difficult to apply point-wise connection structure DL models in critical applications that rely on accurate forecasts, such as medical diagnosis \cite{liao2019clinical} and financial risk assessment \cite{rudin2019stop}. Therefore, enhancing the interpretability of those DL models is crucial to instill confidence in the forecasting process and provide valuable insights into the behavior of time series data.

To address the shortcomings of the above existing studies, this paper introduces a novel time series forecasting network called MPR-Net. Unlike previous approaches that focus mainly on historical series pattern extraction, MPR-Net is a multi-scale hierarchical model consisting of three blocks in each layer: historical multi-scale pattern extraction (HPE) block, multi-scale pattern extension forecasting (PEF) block, and future multi-scale series reconstruction (FSR) block. The HPE block is responsible for extracting valuable historical series patterns for forecasting. It employs a pattern decomposition technique using convolution operations, allowing adaptively adjust convolution kernel parameters through end-to-end training to extract valid patterns. This approach enables the extracted patterns contain trend and periodic information without relying on fixed parameter settings used in methods like sliding average and Fourier transform. Additionally, the HPE block incorporates multivariate information fusion and interaction to leverage relationships among variables. And a correlation attention mechanism is proposed to model variable relationships efficiently. The PEF block takes advantage of the prior knowledge of pattern reproduction, where patterns that occur in the past tend to reappear in the future. It utilizes a pattern reproduction sliding matching approach to identify past patterns related to the current pattern. The subsequent pattern of the identified past pattern is then extended as the future forecast, leveraging the inherent periodicity and trend of the pattern. Compared to existing point-wise connected methods using fully connected layers and attention mechanisms, the PEF block offers an interpretable forecasting process with linear time complexity, presenting a promising paradigm for time series forecasting. The FSR block, similar to the HPE block, reconstructs the predicted future patterns from the PEF block. It also incorporates multivariate information fusion and interaction to obtain the final predicted future series. Finally, the multi-scale hierarchy design of MPR-Net allows the different blocks to decompose and process patterns at various time scales across layers, enabling the identification and utilization of key patterns in historical series that impact future series.

Our main contributions are summarized as follows:

• A new multi-scale hierarchical time series forecasting network (MPR-Net) is constructed that can handle patterns in time series with different temporal scales separately.

•The constructed HPE block can perform adaptive pattern extraction on historical series and accomplish the fusion and interaction of multivariate information in historical series.

•Based on the priori knowledge of pattern reproduction, the constructed PEF block can perform pattern matching and pattern extension on the extracted historical pattern series of the HPE block to obtain the forecasting of future patterns.

•The constructed FSR block can perform adaptive future series reconstruction on the future pattern series and accomplish the fusion and interaction of multivariate information in the future series.

•By taking advantage of the temporal dependencies in the time series, the time complexity of the MPR-Net, which mainly consists of three blocks, is approximately $O(n)$, and its forecasting process is interpretable.

The article is organized as follows. In chapter 2, we review the time series forecasting methods. In chapter 3, the MPR-Net model is introduced. In chapter 4, we conduct experiments on large number of real datasets to verify the advantages and generalization of our method. In chapter 5, we summarize our work.
\section{RELATED WORK}
Time series forecasting techniques have been studied by researchers due to the existence of a wide and important application context. From traditional statistical methods to state-of-the-art deep learning methods, the performance of time series forecasting has improved significantly, but even the current state-of-the-art methods still have shortcomings that need urgent improvement.

Traditional statistical methods, such as the autoregressive (AR) model\cite{yule1927vii}, moving average (MA) model\cite{walker1931periodicity}, and differencing integrated moving average-autoregressive (ARIMA) model\cite{box1970time}, have been widely used for time series forecasting in the past. These methods, constructed based on the assumption of data stationarity and linear causality, are relatively simple to implement and computationally efficient. At the same time, they usually provide interpretable results, allowing analysts to understand underlying patterns and relationships in the data. However, the over-simplicity of the models also leads to the potential difficulty in capturing complex nonlinear patterns and dependencies in time series data, making the forecasting performance unsatisfactory on complex real data sets.

With the development of deep learning techniques, recurrent neural networks (RNN)\cite{elman1990finding}, and its variants such as LSTM \cite{hochreiter1997long}, GRU \cite{cho2014learning} and LSSL \cite{gu2021combining}, have been widely used for time series forecasting. Their ability to capture time-dependence allows more effective features to be extracted from complex realistic data for future forecasting than traditional statistical methods. They can also handle variable-length input sequences, making them suitable for forecasting tasks with different time horizons. However, this cyclic structure can lead to gradient disappearance/explosion situations \cite{pascanu2013difficulty} that hinder the learning of long-term dependencies, making it difficult to maintain accurate information for forecasting over long intervals.

Since the defects of recurrent structure itself are difficult to be solved fundamentally, advanced deep learning methods have gradually used fully connected structures in the forecasting process instead. MLP and transformer methods based on fully connected structures are good at capturing complex nonlinear patterns and long-range time dependencies in time series data, and have become the current mainstream forecasting methods. MLP methods convert the forecasting problem into a regression problem by mapping features extracted from historical series to future series through fully connected neural networks. TCN \cite{bai2018empirical} extracts temporal features of different scales through dilated convolution and causal convolution and gives them to the fully connected layer for forecasting. N-BEATS \cite{oreshkin2019n} and N-HiTS \cite{challu2022n} perform time series decomposition through multiple layers of full connected network, fitting partial information of the time series in each layer.  DLinear \cite{zeng2022transformers} directly uses the fully connected layer to predict the seasonal component and the trend component respectively. TimesNet \cite{wu2022timesnet} maps the input length to the output length through a fully connected layer, and then uses Fourier decomposition and 2D convolution to establish the correlation between different periods for forecasting.

The Transformer, introduced by Google in 2017, make significant advancements in natural language processing (NLP) \cite{kenton2019bert, brown2020language} and has found applications in time series forecasting. Transformer based methods use different attention mechanism to capture global dependencies across time steps, thus allowing them to efficiently model long-term relationships. LogTrans \cite{li2019enhancing} introduces local convolutions into the Transformer and proposes LogSparse attention to select time steps at exponentially increasing intervals to reduce the time complexity. Reformer \cite{kitaev2020reformer} proposes Locality-Sensitive hashing attention instead of self-attention. Informer \cite{zhou2021informer} extends Transformer with ProbSparse attention based on KL-divergence. Non-stationary Transformers \cite{liu2022non} restores the non-stationarity of time series in the computation of attention. Based on periodic trend decomposition, Autoformer \cite{wu2021autoformer} replaces point-to-point correlation calculations by correlation calculations between subsequences; FEDformer \cite{zhou2022fedformer} enhances components of Fourier or wavelet decomposition and performs an attention mechanism based on the components. ETSformer \cite{woo2022etsformer} is based on exponential smoothing and progressively extracts a series of level, growth, and seasonal representations from intermediate latent residues by stacking multiple layers.

Different from the current work, this paper proposes a new type of time series forecasting network MPR-Net, which can utilize the temporal information of the data through the structure of linear time complexity and improve the utilization of information, in addition to constructing an interpretable forecasting model based on the a priori knowledge of pattern reproduction. The leading forecasting performance proves that MPR-Net can provide a new way of thinking for time series forecasting.

\begin{figure*}
	\centering
	\includegraphics[width=190mm]{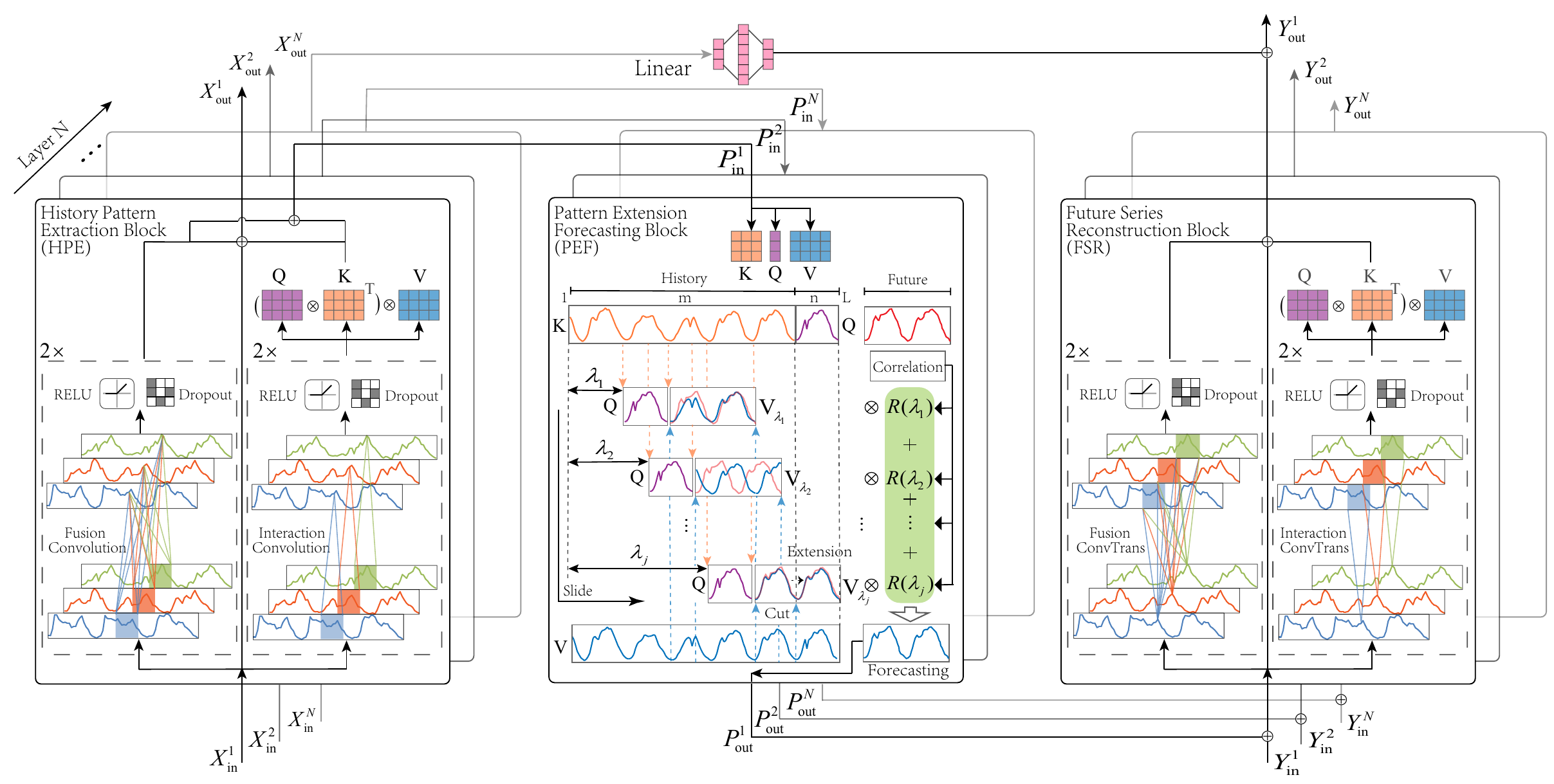}
	\caption{Overview of MPR-Net structure. MPR-Net is a multi-scale hierarchical model consisting of HPE, PEF, and FSR three blocks in each layer. $X_{{\rm{in}}}^1$ and $Y_{{\rm{out}}}^1$ are the model input and output.}
	\label{fig2:env}
\end{figure*}

\section{MPR-Net}
In this section, we provide a detailed description of the proposed MPR-Net, as shown in Fig.1. The construction process of the network is formally formulated through three main modules, while its rationality is elaborated.

\subsection{Overview of the MPR-Net Structure}
The core principle of time series forecasting is that key patterns in historical series will reproduction in the future. MPR-Net exploits this principle by identifying these potential effective patterns within complex historical time series and leveraging them to generate interpretable forecasts. Additionally, MPR-Net tackles the presence of multiple patterns with different time scales commonly found in complex real-time series. It achieves this by employing a multi-scale hierarchical structure, which adaptively decomposes the overlapping and interacting patterns into simpler patterns at various scales.

For a historical time series input $I \in \mathbb{R}^{L \times D}$ of length $L$ and variable dimension $D$, to reduce the difficulty of network training, MPR-Net first performs the normalization process:
\[{I_{{\rm{norm}}}} = I/\sigma (I - {I_{{\rm{mean}}}}),\]
where $I_{{\rm{mean}}}$ denotes the mean value in dimension $L$, and $\sigma$ denotes standard deviation.

As shown in Fig. 1, MPR-Net contains $N$ layers. We first use $X_{{\rm{in}}}^i$ to denote the input of the HPE block at layer $i \in N$ while use $X_{{\rm{out}}}^i$ and $P_{{\rm{in}}}^i$ to denote its outputs. The operation process of the HFE block is represented using the function ${\mathop{\rm HPE}\nolimits} ()$. Among them,

\[X_{{\rm{out}}}^i,P_{{\rm{in}}}^i = {\mathop{\rm HPE}\nolimits} (X_{{\rm{in}}}^i),\]
\[X_{{\rm{in}}}^i = \left\{ {\begin{array}{*{20}{l}}
		{{I_{{\rm{norm}}}},i = 1}\\
		{X_{{\rm{out}}}^{i - 1},i \in (1,N]}
	\end{array}.} \right.\]

Then, $P_{\rm{in}}^i$ is used as input to the PEF block to obtain the output $P_{\rm{out}}^i$, which is represented by the function:
\[P_{{\rm{out}}}^i = {\mathop{\rm PEF}\nolimits} (P_{{\rm{in}}}^i).\]

Next, $P_{{\rm{out}}}^i$ and $Y_{{\rm{in}}}^i$ are added as the input to the FSR block to obtain the output $Y_{\rm{out}}^i$. The operation process of the FSR block is represented using the function ${\mathop{\rm FSR}\nolimits} ()$. Among them,
\[Y_{{\rm{out}}}^i = \left\{ {\begin{array}{*{20}{l}}
		{{\mathop{\rm FSR}\nolimits} (P_{{\rm{out}}}^i + Y_{{\rm{in}}}^i) + {\mathop{\rm Linear}\nolimits} (X_{{\rm{out}}}^N),i = 1}\\
		{{\mathop{\rm FSR}\nolimits} (P_{{\rm{out}}}^i + Y_{{\rm{in}}}^i),i \in (1,N]}
\end{array}} \right.,\]
\[Y_{{\rm{in}}}^i = \left\{ {\begin{array}{*{20}{l}}
		{0,i = N}\\
		{Y_{{\rm{out}}}^{i + 1},i \in [1,N)}
\end{array}} \right.,\]
where ${\mathop{\rm Linear}\nolimits}$ represents a simple linear transformation function, which acts on $X_{{\rm{out}}}^N$ and is used to fit residuals beyond the forecasting information. $Y_{{\rm{out}}}^1$ denotes the future series forecasted by the normalized historical series ${I_{{\rm{norm}}}}$.

The final forecasting series $F \in \mathbb{R}^{T \times D}$ of length $T$ is obtained by inverse normalization of $Y_{\rm{out}}^1$:
\[F = Y_{{\rm{out}}}^1 \times \sigma (I - {I_{{\rm{mean}}}}) + {I_{{\rm{mean}}}}.\]
\subsection{Historical Multi-Scale Pattern Extraction Block}
To forecast future time series, MPR-Net extracts important underlying patterns from complex historical series using the HPE block. This block performs sliding feature extraction adaptively on a complex historical series by means of the learnable convolutional kernel. The extracted feature values can represent the correlation between the historical series and the convolution kernel parameters. Thus, the complex historical series can be decomposed by simple convolution operations into corresponding simple patterns related to convolution parameters. In addition, complex historical series often mixed with different time scales patterns, which increases the difficulty of extracting effective patterns. To solve this problem, MPR-Net constructs a multi-layer structure, which makes HPE block have different temporal scale receptive fields at different layers, so as to carry out adaptive decomposition of patterns at different time scales.

MPR-Net further exploits the fusion and interaction of information from multiple variables in historical series to assist forecasting of future series. The fusion of multivariate information enables the creation of comprehensive feature representations by considering the collective information of each variable. This facilitates a more informative understanding of the sequence state, allowing the model to learn contextual relationships within the historical series. The interaction of multivariate information analyzes and mines the dependencies among variables, establishing correlations between them. Strong correlations indicate that the value of one variable provides valuable information to other variables. By leveraging these correlations, the model becomes more resilient to changes or disturbances in the historical series.

• \textbf{Multivariate Information Fusion Convolution}: With the setup of multiple convolution kernels, HPE blocks in each layer can adaptively extract different patterns in the same temporal scale. And then HPE blocks take multivariate information fusion convolution, which means all channels convolution results form input are added to each new channel in the output. The corresponding process can be formalized as: 
\[F_{i,1}^{{\rm{fusion}}} = {\mathop{\rm Drop}\nolimits} ({\mathop{\rm Relu}\nolimits} ({\mathop{\rm DConv}\nolimits} (X_{{\rm{in}}}^i))),\]
\[F_{i,2}^{{\rm{fusion}}} = {\mathop{\rm Drop}\nolimits} ({\mathop{\rm Relu}\nolimits} ({\mathop{\rm DConv}\nolimits} (F_{i,1}^{{\mathop{\rm fusion}\nolimits} }))),\]
where ${\mathop{\rm DConv}\nolimits} $ denotes dilated convolution to further enlarge the receptive field, and relu activation function and dropout are used to increase the nonlinear expression ability and generalization ability of the model.

• \textbf{Multivariate Information Interaction Convolution}: 
The HPE blocks also utilize multiple convolution kernels, where the number of kernels matches the number of channels (variables). Each kernel convolves independently on its corresponding channel, enabling feature extraction on each channel without influence from other channels. To facilitate the interaction of multivariate information, correlation attention is proposed to the extracted features in the channel dimension. Unlike the traditional attention mechanism, which relies on dot product similarity, the HFE blocks employ the Pearson correlation coefficient to measure the correlation between multivariate features. By considering the correlation, the model can effectively capture the interactions between negatively correlated variables, improving the utilization of time series information. The channel independent convolution process can be represented as follows:

\[F_{i,1}^{{\rm{interact}}} = {\mathop{\rm Drop}\nolimits} ({\mathop{\rm Relu}\nolimits} ({\mathop{\rm IDConv}\nolimits} (X_{{\rm{in}}}^i))),\]
\[F_{i,2}^{{\rm{interact}}} = {\mathop{\rm Drop}\nolimits} ({\mathop{\rm Relu}\nolimits} ({\mathop{\rm IDConv}\nolimits} (F_{i,1}^{{\rm{interact}}}))),\]
where ${\mathop{\rm IDConv}\nolimits} $ denotes channel independent convolution. 

To enhance the expressiveness of the extracted features, $F_{i,2}^{{\rm{interact}}}$ is subjected to three different linear transformations to obtain the feature matrices $Q_{c}$, $K_{c}$ and $V_{c}$. And the correlation attention mechanism can be formalized as:
\[CorAttentio{n_i} = \frac{{Q_{c} \times {K_{c}^{T}}}}{{ d }} \times V_{c},\]
where $d = D \times \sqrt L$ is a scaling factor to prevent the inner product too large.

After end-to-end training, the parameters of the convolution kernel are adaptively adjusted according to the forecasting target, so that the convolution operation can extract effective patterns at different time scales for each variable. Based on these effective patterns, multivariate information can better guide future forecasting through fusion and interaction. The finally output of each HFE block is formalized as:
\[X_{{\rm{out}}}^i = F_{i,2}^{{\rm{fusion}}} + CorAttentio{n_i} + X_{{\rm{in}}}^i,\]
\[P_{{\rm{out}}}^i = F_{i,2}^{{\rm{fusion}}} + CorAttention{n_i},\]
where the operation of adding $X_{\rm{in}}^i$ in $X_{\rm{out}}^i$ means the residual connection to prevent model degeneration in deep network.

\subsection{Multi-Scale Pattern Reproduction Forecasting Block}
Usually the patterns of time series keep shifting over time, and predictable time series exhibit self-similarity of patterns in the time dimension, i.e., patterns that appear in previous series reappear later. Based on this a priori knowledge of pattern reproduction, when the pattern of the current series is similar to the pattern of the past series, the pattern of the future series should also be similar to the subsequent of past series. Therefore, we can use the subsequent of the past series with the same pattern as the current series as a forecasting of the future, a process called pattern extension. In this section, we construct multi-scale PEF blocks to perform future pattern forecasting on the historical patterns extracted by the HPE blocks at the corresponding scales.

\subsubsection{Pattern Reproduction Sliding Matching}
The PEF blocks introduce the Pattern Matching Attention (PMA), an interpretable sequence-level sliding matching attention mechanism. PMA preserves the temporal relationship between data points and matches past patterns similar to the current pattern. This addresses the limitations of the original self-attention mechanism, providing improved computational efficiency and interpretability.

To further enhance the expressiveness of the patterns $P_{\rm{in}}^i$ extracted by the HPE module, the PEF blocks first perform different linear mappings of $P_{\rm{in}}^i$ in different temporal regions to obtain three different lengths matrices of $Q$, $K$ and $V$. Among them, $K\in\mathbb{R}^{m \times D}$ corresponds to the feature mapping of past pattern series, $Q\in\mathbb{R}^{n \times D}$ corresponds to the feature mapping of current pattern series, and $V\in\mathbb{R}^{L \times D}$ corresponds to the feature mapping of the whole historical pattern series. PEF novelly utilizes the priori knowledge of pattern reproduction to construct the PMA structure, which sliding matching of $Q$ in $K$ and using the relevant region in $V$ for future forecasting according to the matching degree. The reason for this setting is that in time series, the current pattern series represented by $Q$ is more closely linked to the future pattern series than other historical pattern series. Therefore, using $Q$ as a query pattern allows finding patterns in the historical pattern series K that are more relevant to future patterns.

Specifically, when the sliding time delay is ${\lambda _j}$ and $j \in [1,m - n)$, the current pattern series $Q$ is used for correlation matching with the previous pattern series ${K_{[{\lambda _j}:{\lambda _j} + n]}}$ in the corresponding region. In order to eliminate the effect of the different scales of the two pattern series values to calculate the correlation coefficient approximatively, PMA subtract the mean values of $Q$ and ${K_{[{\lambda _j}:{\lambda _j} + n]}}$ respectively to make their values of the same scale as shown in Fig.2, thus using the product of $Q$ and ${K_{[{\lambda _j}:{\lambda _j} + n]}}$ as the correlation measure. The correlation of pattern matching can be expressed formally as:
\[R({\lambda _j}) = (Q - {Q_{\rm{mean}}}) \odot ({K_{[{\lambda _j}:{\lambda _j} + n]}} - {K_{[{\lambda _j}:{\lambda _j} + n],\rm{mean}}})\]
where $\odot$ expresses the inner product of elements and sum them up, and $R({\lambda _j})$ denotes the confidence score of correlation matching.

\subsubsection{Forecasting Based On Pattern Extension}
According to the phenomenon that patterns in historical series will be reproduced in the future series, if pattern series $P_{[{\lambda _{L - n}}:{\lambda _L}],in}^i$ is related to pattern series $P_{[{\lambda _j}:{\lambda _{j + n}}],in}^i$, then the future pattern series of $P_{[{\lambda _{L - n}}:{\lambda _L}],in}^i$ should also be related to the subsequent pattern sequences of $P_{[{\lambda _j}:{\lambda _{j + n}}],in}^i$. Thus, when the current pattern series $Q$ is sliding matching on the past pattern series $K$, the PMA cuts the corresponding subsequent pattern series  ${V_{[{\lambda _{j + n + 1}}:L]}}$ on the historical pattern series V as the future forecasting based on the priori knowledge of pattern reproduction. 

When the forecasting length $T$ is longer than the length $C=L-{\lambda _{j + n}}$of the cutted pattern series ${V_{[{\lambda _{j + n + 1}}:L]}}$, the future pattern series lacks the corresponding historical pattern series data for forecasting. To solve this problem, the PEF block adopts pattern extension strategy to fill in the missing data. The pattern extension strategy is a further inference based on the priori knowledge of pattern reproduction: the reproduced pattern will continue to reappear in the future, and the pattern reproduction interval is assumed to be identical based on the periodicity of the data. With this strategy, PMA extends ${V_{[{\lambda _{j + n + 1}}:L]}}$ to the forecasting length by performing multiple splicing, thus making up for the missing data. 

On the other hand, in pattern matching $Q$ and ${K_{[{\lambda _j}:{\lambda _j} + n]}}$ are subtracted from their respective mean values thus avoiding that the correlation calculation is influenced by pattern series with larger values, as shown in Fig.2. However, this approach also results in the loss of information about the relative difference values of the pattern sequences, which can reflect the trend information in the patterns series and can be represented by:
\[{d_{{\lambda _j}}} = {Q_{{\rm{mean}}}} - {K_{[{\lambda _j}:{\lambda _j} + n],{\rm{mean}}}}.\]
Therefore, when using the matched past pattern series ${V_{[{\lambda _{j + n + 1}}:L]}}$ for future forecasting, the relative difference values information $d_{{\lambda _j}}$ needs to be supplemented so that the trend information of the historical pattern series can be continued in the future series.

\begin{figure}
	\centering
	\includegraphics[width=85mm]{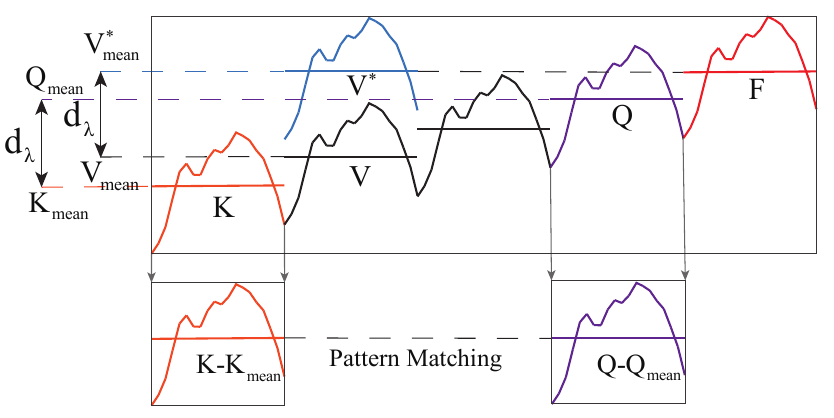}
	\caption{Pattern matching and extension under elimination of scale differences}
	\label{fig2:env}
\end{figure}

In the following we will give a formal representation of the process of pattern extension. First of all, PMA needs to determine the times of pattern extension to be performed by the following equation:
\[a = T/C,b = T\% C,\]
where $/$ means the operation of getting quotient, and $\%$ means the operation of getting remainder. Subsequently, the pattern sequence ${V_{[{\lambda _{j + n + 1}}:L]}}$ is spliced by $a$ times and the information of the relative difference values $d_{{\lambda _j}}$ of the two pattern series is added as a supplement to the trend information during the splicing:

\[V_{{\lambda _j}}^{e\_a} = \sum\limits_{l = 1}^a {({V_{[{\lambda _{j + n + 1}}:L]}} + l \times {d_{{\lambda _j}}}} ),\]
\[V_{{\lambda _j}}^{e\_b} = {V_{[{\lambda _{j + n + 1}}:{\lambda _{j + n + b}}]}} + (a + 1) \times {d_{{\lambda _j}}},\]
\[V_{{\lambda _j}}^* = V_{{\lambda _j}}^{e\_a} \oplus V_{{\lambda _j}}^{e\_b},\]
where $\sum $ and $\oplus$ are mean the operation of splicing, $V_{{\lambda _j}}^*$ means finally forecasting data getting from pattern extension and fused with relative difference values $d_{{\lambda _j}}$.

By the above way, PMA obtains the corresponding pattern forecasting data $V_{{\lambda _j}}^e$ at different time delays. The final forecasting of future patterns will take into account the correlation ${R({\lambda _j})}$ of pattern matching at different time delays and the forecasting data $V_{{\lambda _j}}^*$ of the corresponding pattern extension, which can be expressed by:
\[P_{\rm{out}}^i = \sum\limits_{j = 1}^{m - n} {\frac{{R({\lambda _j}) \times V_{{\lambda _j}}^*}}{d}}, \]
where $d = L \times \sqrt D$ is a scaling factor to prevent the inner product too large.


\subsection{Future Multi-Scale Series Reconstruction Block}
Finally, MPR-net constructs the FSR blocks to reconstruct the forecasting future pattern series at different scales to recover to future series. And the final future series forecasting based on pattern reproduction will be obtained by fusion of these reconstructed series.

The overall structure of the FFR module is similar to that of HFE in that it also contains the fusion and interaction of multivariate information, except that it is geared towards processing future pattern seriies. The difference is that in the pattern extraction prior to the fusion and interaction of multiple variables, the FFR blocks use the transposed convolution operation opposite to the convolution operation in the HFE blocks, thus reconstructing the convolutionally extracted pattern feature back to the original series level.

• \textbf{Multivariate Information Fusion Transposed Convolution}: The process of reconstructing multivariate future pattern series and fusing multivariate information can be formulated as:
\[F_{i,1}^{{\rm{fusion}}} = {\mathop{\rm Drop}\nolimits} ({\mathop{\rm Relu}\nolimits} ({\mathop{\rm TConv}\nolimits} (Y_{{\rm{in}}}^i + P_{{\rm{out}}}^i))),\]
\[F_{i,2}^{{\rm{fusion}}} = {\mathop{\rm Drop}\nolimits} ({\mathop{\rm Relu}\nolimits} ({\mathop{\rm TConv}\nolimits} (F_{i,1}^{{\rm{fusion}}}))),\]
where $TConv$ means the operation of transposed convolution.

• \textbf{Multivariate Information Interaction Convolution}: The process of reconstructing multivariate future feature series and interacting multivariate information can be formulated as:
\[F_{i,1}^{{\rm{interact}}} = {\mathop{\rm Drop}\nolimits} ({\mathop{\rm Relu}\nolimits} ({\mathop{\rm ITConv}\nolimits} (Y_{{\rm{in}}}^i + P_{{\rm{out}}}^i))),\]
\[F_{i,2}^{{\rm{interact}}} = {\mathop{\rm Drop}\nolimits} ({\mathop{\rm Relu}\nolimits} ({\mathop{\rm ITConv}\nolimits} (F_{i,1}^{{\rm{interact}}}))),\]
where $\rm{ITConv}$ denotes the operation of channel independent transposed convolution. Also to enhance the expression of the features, FFR blocks perform three different linear transformations on the feature series extracted by transposed convolution to obtain ${Q_{tc}}$, ${K_{tc}}$ and ${V_{tc}}$. They are then subjected to the calculation of correlation attention:
\[CorAttentio{n_i} = \frac{{{Q_{tc}} \times K_{tc}^T}}{{ d }} \times {V_{tc}},\]
where $d = D \times \sqrt T$ is a scaling factor to prevent the inner product too large.

\begin{figure}
	\centering
	\includegraphics[width=80mm]{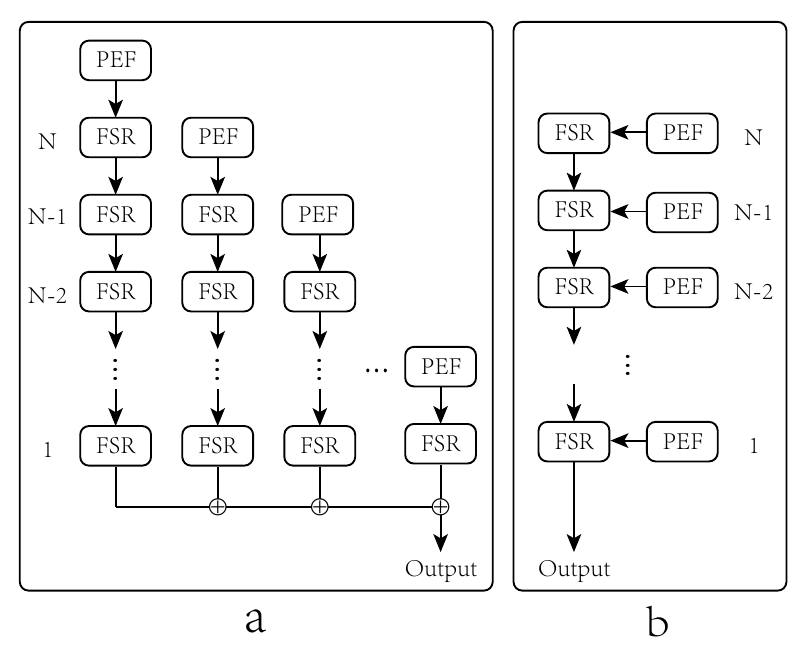}
	\caption{Predicted future patterns reconstructed by FSR block with shared structure.}
	\label{fig2:env}
\end{figure}

The FSR block and the HRE block also have a key difference in their operation order at different layers. In the HPE block, lower-level block outputs serve as inputs to higher-level blocks, enabling pattern extraction at larger scales. However, when reconstructing the forecasted patterns from the PEF block at the $N$-th layer using the FSR block at the same layer, the feature level of the pattern is only recovered to the $N-1$-th layer. Therefore, the output of the PEF block of each layer needs to be reconstructed by the FSR block for the corresponding number of times, as shown in Fig.3-a. In order to avoid a large number of FSR blocks operations, MPR-Net introduces a shared structure, where the future patterns predicted at different PEF block layers can be reconstructed by the same FSR block at a lower feature level, as shown in Fig.3-b. This shared structure reduces the complexity of the model and improves running speed. In this structure, the input of each layer of the PEF block is $Y_{in}^i + P_{out}^i$, and the output is:

\[Y_{{\rm{out}}}^i = F_{i,2}^{{\rm{fusion}}} + CorAttentio{n_i} + Y_{{\rm{in}}}^i + P_{{\rm{out}}}^i,\]
where the operation of adding $Y_{in}^i + P_{out}^i$ means the residual connection to prevent model degeneration in deep network. In the first layer, we further add $\rm{Linear}(X_{{\rm{out}}}^N)$ to $Y_{{\rm{out}}}^1$ as a fit of residual information.

\subsection{Algorithm Flow and Time Complexity Analysis}
In the above sections, we have described in detail the entire framework of MPR-Net and its three blocks: HFE, PEF, and FFR. To give an overview of the model process, Algorithm 1 constructs an algorithm flowchart to give a concise overview of the flow of the MPR-Net method. 

\begin{algorithm}
	\caption{MPR-Net}
	\begin{algorithmic}
		\REQUIRE $I \in \mathbb{R}^{L \times D}$, $N \ge 1$                 
		\ENSURE $F \in \mathbb{R}^{T \times D}$     
		\STATE $X_{{\rm{in}}}^1 = I/\sigma (I - {I_{{\rm{mean}}}})$
		\FOR {$i = 1$ to $N$} 
		\STATE $X_{{\rm{out}}}^i,P_{{\rm{in}}}^i = {\mathop{\rm HPE}\nolimits} (X_{{\rm{in}}}^i)$
		\STATE $P_{\rm{out}}^i = \rm{PEF}(P_{\rm{in}}^i)$
		\STATE $X_{{\rm{in}}}^{i + 1} = X_{{\rm{out}}}^i$
		\ENDFOR
		\STATE $Y_{{\rm{in}}}^N = 0$
		\FOR {$i = N$ to $1$}
		\IF {$i > 1$}                        
		\STATE $Y_{{\rm{out}}}^i = {\mathop{\rm FSR}\nolimits} (P_{{\rm{out}}}^i + Y_{{\rm{in}}}^i)$               
		\ELSE
		\STATE $Y_{{\rm{out}}}^i = {\mathop{\rm FSR}\nolimits} (P_{{\rm{out}}}^i + Y_{{\rm{in}}}^i) + {\mathop{\rm Linear}\nolimits} (X_{{\rm{out}}}^N)$
		\ENDIF 
		\ENDFOR
		\STATE $F = Y_{out}^1 \times \sigma (I - {I_{{\rm{mean}}}}) + {I_{mean}}$
		\RETURN $F$
	\end{algorithmic}
\end{algorithm}

\section{Experiment}
To validate the predictive performance of the proposed MPR-Net, we have conducted adequate experiments on several real datasets and performed detailed analysis of the results. This section first describes the adopted datasets, evaluation metrics, comparison methods and experimental setup. Then, the proposed method is compared in multiple experiments from three aspects: 1. forecasting performance under long-term and short-term forecasting objectives; 2. the effect of model structure on forecasting performance; and 3. the effect of adding additional information on forecasting performance.

\begin{table}[]
	\caption{Datasets used for long and short term forecasting tasks}
	\resizebox{82mm}{28mm}{
		\renewcommand{\arraystretch}{1.5}
		\begin{tabular}{|c|c|c|c|}
			\hline
			Dataset      & Dim & Dataset Size         & Information            \\ \hline
			ETTm1,ETTm2  & 7   & (34465,11521,11521)  & Electricity(15mins)    \\ \hline
			ETTh1,ETTh2  & 7   & (8545, 2881, 2881)   & Electricity(15mins)    \\ \hline
			Electricity  & 321 & (18317, 2633, 5261)  & Electricity(Hourly)    \\ \hline
			Traffic      & 862 & (12185, 1757, 3509)  & Transportation(Hourly) \\ \hline
			Weather      & 21  & (36792, 5271, 10540) & Weather(10mins)        \\ \hline
			Exchange     & 8   & (5120, 665, 1422)    & Exchange rate(Daily)   \\ \hline
			ILI          & 7   & (617, 74, 170)       & Illness(Weekly)        \\ \hline
			M4-Yearly    & 1   & (23000,0,23000)      & Demographic            \\ \cline{1-3}
			M4-Quarterly & 1   & (24000,0,24000)      & Finance                \\ \cline{1-3}
			M4-Monthly   & 1   & (48000,0,48000)      & Industry               \\ \cline{1-3}
			M4-Weakly    & 1   & (359,0,359)          & Macro                  \\ \cline{1-3}
			M4-Daily     & 1   & (4227,0,4227)        & Micro                  \\ \cline{1-3}
			M4-Hourly    & 1   & (414,0,414)          & Other                  \\ \hline
	\end{tabular}}
\end{table}

\subsection{Dataset}
To verify the robustness of MPR-Net forecasting performance, we selected multiple datasets(Table 1) for experiments on two mainstream forecasting tasks (long-term forecasting and short-term forecasting). 

The data sets used for the long-term forecasting are:
(1)Electric Transformer Temperature\cite{zhou2021informer}: It contains four sub-data sets (ETTm1, ETTm2, ETTh1, ETTh2) that collect data about the load and oil temperature of power transformers.
(2)Electricity\footnote{https://archive.ics.uci.edu/ml/datasets/ElectricityLoadDiagrams 20112014}: The electricity consumption of 321 customers are recorded.
(3)Traffic\footnote{http://pems.dot.ca.gov}: Road occupancy in the California Highway system recorded by sensors
(4)Weather\footnote{https://www.bgc-jena.mpg.de/wetter/}: 21 meteorological indicators for Germany over one year period are included.
(5)Exchange\cite{lai2018modeling}: Contains daily exchange rates for eight countries
(6)ILI\footnote{https://gis.cdc.gov/grasp/fluview/fluportaldashboard.html}: The percentage of patients with influenza was recorded from data collected by the U.S. Centers for Disease Control and Prevention.

The data set used for the short-term forecasting is M4 \footnote{https://github.com/Mcompetitions/M4-methods}, which involve 6 subsets of collected univariate marketing data: M4-Yearly, M4-Quarterly, M4-Monthly, M4-Weakly, M4-Daily, M4-Hourly.

\begin{table*}[!ht]
	\renewcommand{\arraystretch}{1.2}
	\centering
	\caption{Complete results for the long-term forecasting task. A wide range of advanced models are compared at different forecasting lengths. The input series length is set to 36 for the ILI dataset and 96 for the other datasets. Avg is the average of all four forecasting lengths.}
	\resizebox{\textwidth}{60mm}{
		\begin{tabular}{c c c c|c c|c c|c c|c c|c c|c c|c c|c c|c c|c c|c c|c c}
			\hline
			\multicolumn{2}{c}{\multirow{2}{*}{Models}}&\multicolumn{2}{c}{\textbf{MPR-Net}}&\multicolumn{2}{c}{TimesNet}&\multicolumn{2}{c}{ETSformer}&\multicolumn{2}{c}{LightTS}&\multicolumn{2}{c}{DLinear}&\multicolumn{2}{c}{FEDformer}&\multicolumn{2}{c}{Stationary}&\multicolumn{2}{c}{Autoformer}&\multicolumn{2}{c}{Pyraformer}&\multicolumn{2}{c}{Informer}&\multicolumn{2}{c}{LogTrans}&\multicolumn{2}{c}{Reformer}&\multicolumn{2}{c}{LSSL}\\
			&&\multicolumn{2}{c}{\textbf{(Ours)}}&\multicolumn{2}{c}{(2023)}&\multicolumn{2}{c}{(2022)}&\multicolumn{2}{c}{(2022)}&\multicolumn{2}{c}{(2023)}&\multicolumn{2}{c}{(2022)}&\multicolumn{2}{c}{(2022a)}&\multicolumn{2}{c}{(2021)}&\multicolumn{2}{c}{(2021a)}&\multicolumn{2}{c}{(2021)}&\multicolumn{2}{c}{(2019)}&\multicolumn{2}{c}{(2020)}&\multicolumn{2}{c}{(2022)}\\
			\cline{3-28}
			\multicolumn{2}{c}{Metric}&MSE&MAE&MSE&MAE&MSE&MAE&MSE&MAE&MSE&MAE&MSE&MAE&MSE&MAE&MSE&MAE&MSE&MAE&MSE&MAE&MSE&MAE&MSE&MAE&MSE&MAE\\
			\hline
			\multicolumn{1}{c|}{\multirow{5}{*}{\rotatebox{90}{ETTm1}}}&\multicolumn{1}{c|}{96}&\color{red}{\textbf{0.313}}&\color{red}{\textbf{0.355}}&\color{blue}{\textbf{0.338}}&0.375&0.375&0.398&0.374&0.400&0.345&\color{blue}{\textbf{0.372}}&0.379&0.419&0.386&0.398&0.505&0.475&0.543&0.510&0.672&0.571&0.600&0.546&0.538&0.528&0.450&0.477\\
			\multicolumn{1}{c|}{}&\multicolumn{1}{c|}{192}&\color{red}{\textbf{0.361}}&\color{red}{\textbf{0.382}}&\color{blue}{\textbf{0.374}}&\color{blue}{\textbf{0.387}}&0.408&0.410&0.400&0.407&0.380&0.389&0.426&0.441&0.459&0.444&0.553&0.496&0.557&0.537&0.795&0.669&0.837&0.700&0.658&0.592&0.469&0.481\\
			\multicolumn{1}{c|}{}&\multicolumn{1}{c|}{336}&\color{red}{\textbf{0.392}}&\color{red}{\textbf{0.403}}&\color{blue}{\textbf{0.410}}&\color{blue}{\textbf{0.411}}&0.435&0.428&0.438&0.438&0.413&0.413&0.445&0.459&0.495&0.464&0.621&0.537&0.754&0.655&1.212&0.871&1.124&0.832&0.898&0.721&0.583&0.574\\
			\multicolumn{1}{c|}{}&\multicolumn{1}{c|}{720}&\color{red}{\textbf{0.459}}&\color{red}{\textbf{0.441}}&0.478&\color{blue}{\textbf{0.450}}&0.499&0.462&0.527&0.502&\color{blue}{\textbf{0.474}}&0.453&0.543&0.490&0.585&0.516&0.671&0.561&0.908&0.724&1.166&0.823&1.153&0.820&1.102&0.841&0.632&0.596\\
			\cline{2-28}
			\multicolumn{1}{c|}{}&\multicolumn{1}{c|}{Avg}&\color{red}{\textbf{0.381}}&\color{red}{\textbf{0.395}}&\color{blue}{\textbf{0.400}}&\color{blue}{\textbf{0.406}}&0.429&0.425&0.435&0.437&0.403&0.407&0.448&0.452&0.481&0.456&0.588&0.517&0.691&0.607&0.961&0.734&0.929&0.725&0.799&0.671&0.533&0.532\\
			\hline
			\multicolumn{1}{c|}{\multirow{5}{*}{\rotatebox{90}{ETTm2}}}&\multicolumn{1}{c|}{96}&\color{red}{\textbf{0.172}}&\color{red}{\textbf{0.255}}&\color{blue}{\textbf{0.187}}&\color{blue}{\textbf{0.267}}&0.189&0.280&0.209&0.308&0.193&0.292&0.203&0.287&0.192&0.274&0.255&0.339&0.435&0.507&0.365&0.453&0.768&0.642&0.658&0.619&0.243&0.342\\
			\multicolumn{1}{c|}{}&\multicolumn{1}{c|}{192}&\color{red}{\textbf{0.238}}&\color{red}{\textbf{0.298}}&\color{blue}{\textbf{0.249}}&\color{blue}{\textbf{0.309}}&0.253&0.319&0.311&0.382&0.284&0.362&0.269&0.328&0.280&0.339&0.281&0.340&0.730&0.673&0.533&0.563&0.989&0.757&1.078&0.827&0.392&0.448\\
			\multicolumn{1}{c|}{}&\multicolumn{1}{c|}{336}&\color{red}{\textbf{0.297}}&\color{red}{\textbf{0.337}}&0.321&\color{blue}{\textbf{0.351}}&\color{blue}{\textbf{0.314}}&0.357&0.442&0.466&0.369&0.427&0.325&0.366&0.334&0.361&0.339&0.372&1.201&0.845&1.363&0.887&1.334&0.872&1.549&0.972&0.932&0.724\\
			\multicolumn{1}{c|}{}&\multicolumn{1}{c|}{720}&\color{red}{\textbf{0.399}}&\color{red}{\textbf{0.397}}&\color{blue}{\textbf{0.408}}&\color{blue}{\textbf{0.403}}&0.414&0.413&0.675&0.587&0.554&0.522&0.421&0.415&0.417&0.413&0.433&0.432&3.625&1.451&3.379&1.338&3.048&1.328&2.631&1.242&1.372&0.879\\
			\cline{2-28}
			\multicolumn{1}{c|}{}&\multicolumn{1}{c|}{Avg}&\color{red}{\textbf{0.277}}&\color{red}{\textbf{0.322}}&\color{blue}{\textbf{0.291}}&\color{blue}{\textbf{0.333}}&0.293&0.342&0.409&0.436&0.350&0.401&0.305&0.349&0.306&0.347&0.327&0.371&1.498&0.869&1.410&0.810&1.535&0.900&1.479&0.915&0.735&0.598\\
			\hline
			\multicolumn{1}{c|}{\multirow{5}{*}{\rotatebox{90}{ETTh1}}}&\multicolumn{1}{c|}{96}&\color{blue}{\textbf{0.383}}&\color{red}{\textbf{0.398}}&0.384&0.402&0.494&0.479&0.424&0.432&0.386&\color{blue}{\textbf{0.400}}&\color{red}{\textbf{0.376}}&0.419&0.513&0.491&0.449&0.459&0.664&0.612&0.865&0.713&0.878&0.740&0.837&0.728&0.548&0.528\\
			\multicolumn{1}{c|}{}&\multicolumn{1}{c|}{192}&\color{blue}{\textbf{0.435}}&\color{red}{\textbf{0.427}}&0.436&\color{blue}{\textbf{0.429}}&0.538&0.504&0.475&0.462&0.437&0.432&\color{red}{\textbf{0.420}}&0.448&0.534&0.504&0.500&0.482&0.790&0.681&1.008&0.792&1.037&0.824&0.923&0.766&0.542&0.526\\
			\multicolumn{1}{c|}{}&\multicolumn{1}{c|}{336}&\color{blue}{\textbf{0.475}}&\color{red}{\textbf{0.446}}&0.491&0.469&0.574&0.521&0.518&0.488&0.481&\color{blue}{\textbf{0.459}}&\color{red}{\textbf{0.459}}&0.465&0.588&0.535&0.521&0.496&0.891&0.738&1.107&0.809&1.238&0.932&1.097&0.835&1.298&0.942\\
			\multicolumn{1}{c|}{}&\multicolumn{1}{c|}{720}&\color{red}{\textbf{0.490}}&\color{red}{\textbf{0.471}}&0.521&\color{blue}{\textbf{0.500}}&0.562&0.535&0.547&0.533&0.519&0.516&\color{blue}{\textbf{0.506}}&0.507&0.643&0.616&0.514&0.512&0.963&0.782&1.181&0.865&1.135&0.852&1.257&0.889&0.721&0.659\\
			\cline{2-28}
			\multicolumn{1}{c|}{}&\multicolumn{1}{c|}{Avg}&\color{blue}{\textbf{0.446}}&\color{red}{\textbf{0.436}}&0.458&\color{blue}{\textbf{0.450}}&0.542&0.510&0.491&0.479&0.456&0.452&\color{red}{\textbf{0.440}}&0.460&0.570&0.537&0.496&0.487&0.827&0.703&1.040&0.795&1.072&0.837&1.029&0.805&0.777&0.664\\
			\hline
			\multicolumn{1}{c|}{\multirow{5}{*}{\rotatebox{90}{ETTh2}}}&\multicolumn{1}{c|}{96}&\color{red}{\textbf{0.289}}&\color{red}{\textbf{0.340}}&0.340&0.374&0.340&0.391&0.397&0.437&\color{blue}{\textbf{0.333}}&\color{blue}{\textbf{0.387}}&0.358&0.397&0.476&0.458&0.346&0.388&0.645&0.597&3.755&1.525&2.116&1.197&2.626&1.317&1.616&1.036\\
			\multicolumn{1}{c|}{}&\multicolumn{1}{c|}{192}&\color{red}{\textbf{0.375}}&\color{red}{\textbf{0.392}}&\color{blue}{\textbf{0.402}}&\color{blue}{\textbf{0.414}}&0.430&0.439&0.520&0.504&0.477&0.476&0.429&0.439&0.512&0.493&0.456&0.452&0.788&0.683&5.602&1.931&4.315&1.635&11.12&2.979&2.083&1.197\\
			\multicolumn{1}{c|}{}&\multicolumn{1}{c|}{336}&\color{red}{\textbf{0.416}}&\color{red}{\textbf{0.425}}&\color{blue}{\textbf{0.452}}&\color{blue}{\textbf{0.452}}&0.485&0.479&0.626&0.559&0.594&0.541&0.496&0.487&0.552&0.551&0.482&0.486&0.907&0.747&4.721&1.835&1.124&1.604&9.323&2.769&2.970&1.439\\
			\multicolumn{1}{c|}{}&\multicolumn{1}{c|}{720}&\color{red}{\textbf{0.424}}&\color{red}{\textbf{0.439}}&\color{blue}{\textbf{0.462}}&\color{blue}{\textbf{0.468}}&0.500&0.497&0.863&0.672&0.831&0.657&0.463&0.474&0.562&0.560&0.515&0.511&0.963&0.783&3.647&1.625&3.188&1.540&3.874&1.697&2.576&1.363\\
			\cline{2-28}
			\multicolumn{1}{c|}{}&\multicolumn{1}{c|}{Avg}&\color{red}{\textbf{0.376}}&\color{red}{\textbf{0.399}}&\color{blue}{\textbf{0.414}}&\color{blue}{\textbf{0.427}}&0.439&0.452&0.602&0.543&0.559&0.515&0.437&0.449&0.526&0.516&0.450&0.459&0.826&0.703&4.431&1.729&2.686&1.494&6.736&2.191&2.311&1.259\\
			\hline
			\multicolumn{1}{c|}{\multirow{5}{*}{\rotatebox{90}{Electricity}}}&\multicolumn{1}{c|}{96}&\color{red}{\textbf{0.142}}&\color{red}{\textbf{0.240}}&\color{blue}{\textbf{0.168}}&\color{blue}{\textbf{0.272}}&0.187&0.304&0.207&0.307&0.197&0.282&0.193&0.308&0.169&0.273&0.201&0.317&0.386&0.449&0.274&0.368&0.258&0.357&0.312&0.402&0.300&0.392\\
			\multicolumn{1}{c|}{}&\multicolumn{1}{c|}{192}&\color{red}{\textbf{0.157}}&\color{red}{\textbf{0.255}}&0.184&0.289&0.199&0.315&0.213&0.316&0.196&\color{blue}{\textbf{0.285}}&0.201&0.315&\color{blue}{\textbf{0.182}}&0.286&0.222&0.334&0.378&0.443&0.296&0.386&0.266&0.368&0.348&0.433&0.297&0.390\\
			\multicolumn{1}{c|}{}&\multicolumn{1}{c|}{336}&\color{red}{\textbf{0.170}}&\color{red}{\textbf{0.269}}&\color{blue}{\textbf{0.198}}&\color{blue}{\textbf{0.300}}&0.212&0.329&0.230&0.333&0.209&0.301&0.214&0.329&0.200&0.304&0.231&0.338&0.376&0.443&0.300&0.394&0.280&0.380&0.350&0.433&0.317&0.403\\
			\multicolumn{1}{c|}{}&\multicolumn{1}{c|}{720}&\color{red}{\textbf{0.197}}&\color{red}{\textbf{0.292}}&\color{blue}{\textbf{0.220}}&\color{blue}{\textbf{0.320}}&0.233&0.345&0.265&0.360&0.245&0.333&0.246&0.355&0.222&0.321&0.254&0.361&0.376&0.445&0.373&0.439&0.283&0.376&0.340&0.420&0.338&0.417\\
			\cline{2-28}
			\multicolumn{1}{c|}{}&\multicolumn{1}{c|}{Avg}&\color{red}{\textbf{0.167}}&\color{red}{\textbf{0.264}}&\color{blue}{\textbf{0.192}}&\color{blue}{\textbf{0.295}}&0.208&0.323&0.229&0.329&0.212&0.300&0.214&0.327&0.193&0.296&0.227&0.338&0.379&0.445&0.311&0.397&0.272&0.370&0.338&0.422&0.313&0.401\\
			\hline
			\multicolumn{1}{c|}{\multirow{5}{*}{\rotatebox{90}{Traffic}}}&\multicolumn{1}{c|}{96}&\color{red}{\textbf{0.429}}&\color{red}{\textbf{0.285}}&0.593&\color{blue}{\textbf{0.321}}&0.607&0.392&0.615&0.391&0.650&0.396&\color{blue}{\textbf{0.587}}&0.366&0.612&0.338&0.613&0.388&0.867&0.468&0.719&0.391&0.684&0.384&0.732&0.423&0.798&0.436\\
			\multicolumn{1}{c|}{}&\multicolumn{1}{c|}{192}&\color{red}{\textbf{0.450}}&\color{red}{\textbf{0.301}}&0.617&\color{blue}{\textbf{0.336}}&0.621&0.399&0.601&0.382&\color{blue}{\textbf{0.598}}&0.370&0.604&0.373&0.613&0.340&0.616&0.382&0.869&0.467&0.696&0.379&0.685&0.390&0.733&0.420&0.849&0.481\\
			\multicolumn{1}{c|}{}&\multicolumn{1}{c|}{336}&\color{red}{\textbf{0.467}}&\color{red}{\textbf{0.315}}&0.629&0.336&0.622&0.396&0.613&0.386&\color{blue}{\textbf{0.605}}&0.373&0.621&0.383&0.618&\color{blue}{\textbf{0.328}}&0.622&0.337&0.881&0.469&0.777&0.420&0.734&0.408&0.742&0.420&0.828&0.476\\
			\multicolumn{1}{c|}{}&\multicolumn{1}{c|}{720}&\color{red}{\textbf{0.508}}&\color{red}{\textbf{0.337}}&0.640&\color{blue}{\textbf{0.350}}&0.632&0.396&0.658&0.407&0.645&0.394&\color{blue}{\textbf{0.626}}&0.382&0.653&0.355&0.660&0.408&0.896&0.473&0.864&0.472&0.717&0.396&0.755&0.423&0.854&0.489\\
			\cline{2-28}
			\multicolumn{1}{c|}{}&\multicolumn{1}{c|}{Avg}&\color{red}{\textbf{0.464}}&\color{red}{\textbf{0.310}}&0.620&\color{blue}{\textbf{0.336}}&0.621&0.396&0.622&0.392&0.625&0.383&\color{blue}{\textbf{0.610}}&0.376&0.624&0.340&0.628&0.379&0.878&0.469&0.764&0.416&0.705&0.395&0.741&0.422&0.832&0.471\\
			\hline
			\multicolumn{1}{c|}{\multirow{5}{*}{\rotatebox{90}{Weather}}}&\multicolumn{1}{c|}{96}&\color{red}{\textbf{0.165}}&\color{red}{\textbf{0.213}}&\color{blue}{\textbf{0.172}}&\color{blue}{\textbf{0.220}}&0.197&0.281&0.182&0.242&0.196&0.255&0.217&0.296&0.173&0.223&0.266&0.336&0.622&0.556&0.300&0.384&0.458&0.490&0.689&0.596&0.174&0.252\\
			\multicolumn{1}{c|}{}&\multicolumn{1}{c|}{192}&\color{red}{\textbf{0.211}}&\color{red}{\textbf{0.253}}&\color{blue}{\textbf{0.219}}&\color{blue}{\textbf{0.261}}&0.237&0.312&0.227&0.287&0.237&0.296&0.276&0.336&0.245&0.285&0.307&0.367&0.739&0.624&0.598&0.544&0.658&0.589&0.752&0.638&0.238&0.313\\
			\multicolumn{1}{c|}{}&\multicolumn{1}{c|}{336}&\color{red}{\textbf{0.269}}&\color{red}{\textbf{0.294}}&\color{blue}{\textbf{0.280}}&\color{blue}{\textbf{0.306}}&0.298&0.353&0.282&0.334&0.283&0.335&0.339&0.380&0.321&0.338&0.359&0.395&1.004&0.753&0.578&0.523&0.797&0.652&0.639&0.596&0.287&0.355\\
			\multicolumn{1}{c|}{}&\multicolumn{1}{c|}{720}&\color{red}{\textbf{0.344}}&\color{red}{\textbf{0.343}}&0.365&\color{blue}{\textbf{0.359}}&0.352&0.288&0.352&0.386&\color{blue}{\textbf{0.345}}&0.381&0.403&0.428&0.414&0.410&0.419&0.428&1.420&0.934&1.059&0.741&0.869&0.675&1.130&0.792&0.384&0.415\\
			\cline{2-28}
			\multicolumn{1}{c|}{}&\multicolumn{1}{c|}{Avg}&\color{red}{\textbf{0.247}}&\color{red}{\textbf{0.276}}&\color{blue}{\textbf{0.259}}&\color{blue}{\textbf{0.287}}&0.271&0.334&0.261&0.312&0.265&0.317&0.309&0.360&0.288&0.314&0.338&0.382&0.946&0.717&0.634&0.548&0.696&0.602&0.803&0.656&0.271&0.334\\
			\hline
			\multicolumn{1}{c|}{\multirow{5}{*}{\rotatebox{90}{Exchange}}}&\multicolumn{1}{c|}{96}&\color{blue}{\textbf{0.087}}&\color{blue}{\textbf{0.206}}&0.107&0.234&\color{red}{\textbf{0.085}}&\color{red}{\textbf{0.204}}&0.116&0.262&0.088&0.218&0.148&0.278&0.111&0.237&0.197&0.323&1.748&1.105&0.847&0.752&0.968&0.812&1.065&0.829&0.395&0.474\\
			\multicolumn{1}{c|}{}&\multicolumn{1}{c|}{192}&\color{blue}{\textbf{0.182}}&\color{blue}{\textbf{0.304}}&0.226&0.344&\color{blue}{\textbf{0.182}}&\color{red}{\textbf{0.303}}&0.215&0.359&\color{red}{\textbf{0.176}}&0.315&0.271&0.380&0.219&0.335&0.300&0.369&1.874&1.151&1.204&0.895&1.040&0.851&1.188&0.906&0.776&0.698\\
			\multicolumn{1}{c|}{}&\multicolumn{1}{c|}{336}&\color{blue}{\textbf{0.337}}&\color{red}{\textbf{0.418}}&0.367&0.448&0.348&0.428&0.377&0.466&\color{red}{\textbf{0.313}}&\color{blue}{\textbf{0.427}}&0.460&0.500&0.421&0.476&0.509&0.524&1.943&1.172&1.672&1.036&1.659&1.081&1.357&0.976&1.029&0.797\\
			\multicolumn{1}{c|}{}&\multicolumn{1}{c|}{720}&0.886&0.710&0.964&0.746&1.025&0.774&\color{red}{\textbf{0.831}}&\color{blue}{\textbf{0.699}}&\color{blue}{\textbf{0.839}}&\color{red}{\textbf{0.695}}&1.195&0.841&1.092&0.769&1.447&0.941&2.085&1.206&2.478&1.310&1.941&1.127&1.510&1.016&2.283&1.222\\
			\cline{2-28}
			\multicolumn{1}{c|}{}&\multicolumn{1}{c|}{Avg}&\color{blue}{\textbf{0.373}}&\color{red}{\textbf{0.410}}&0.416&0.443&0.410&0.427&0.385&0.447&\color{red}{\textbf{0.354}}&\color{blue}{\textbf{0.414}}&0.519&0.500&0.461&0.454&0.613&0.539&1.913&1.159&1.550&0.998&1.402&0.968&1.280&0.932&1.121&0.798\\
			\hline
			\multicolumn{1}{c|}{\multirow{5}{*}{\rotatebox{90}{ILI}}}&\multicolumn{1}{c|}{24}&\color{red}{\textbf{1.860}}&\color{red}{\textbf{0.841}}&2.317&\color{blue}{\textbf{0.934}}&2.527&1.020&8.313&2.144&2.398&1.040&3.228&1.260&\color{blue}{\textbf{2.294}}&0.945&3.483&1.287&7.394&2.012&5.764&1.677&4.480&1.444&4.400&1.382&4.381&1.425\\
			\multicolumn{1}{c|}{}&\multicolumn{1}{c|}{36}&\color{blue}{\textbf{1.876}}&\color{blue}{\textbf{0.856}}&1.972&0.920&2.615&1.007&6.631&1.902&2.646&1.088&2.679&1.080&\color{red}{\textbf{1.825}}&\color{red}{\textbf{0.848}}&3.103&1.148&7.551&2.031&4.755&1.467&4.799&1.467&4.783&1.448&4.442&1.416\\
			\multicolumn{1}{c|}{}&\multicolumn{1}{c|}{48}&\color{red}{\textbf{1.947}}&\color{red}{\textbf{0.861}}&2.238&0.940&2.359&0.972&7.299&1.982&2.614&1.086&2.622&1.078&\color{blue}{\textbf{2.010}}&\color{blue}{\textbf{0.900}}&2.669&1.085&7.662&2.057&4.763&1.469&4.800&1.468&4.832&1.465&4.559&1.443\\
			\multicolumn{1}{c|}{}&\multicolumn{1}{c|}{60}&\color{red}{\textbf{1.936}}&\color{red}{\textbf{0.885}}&\color{blue}{\textbf{2.027}}&\color{blue}{\textbf{0.928}}&2.487&1.016&7.283&1.985&2.804&1.146&2.857&1.157&2.178&0.963&2.770&1.125&7.931&2.100&5.264&1.564&5.278&1.560&4.882&1.483&4.651&1.474\\
			\cline{2-28}
			\multicolumn{1}{c|}{}&\multicolumn{1}{c|}{Avg}&\color{red}{\textbf{1.905}}&\color{red}{\textbf{0.861}}&2.139&0.931&2.497&1.004&7.382&2.003&2.616&1.090&2.847&1.144&\color{blue}{\textbf{2.077}}&\color{blue}{\textbf{0.914}}&3.006&1.161&7.635&2.050&5.137&1.544&4.839&1.485&4.724&1.445&4.508&1.440\\
			\hline
			\multicolumn{2}{c}{$1^{st}$ Count}&\multicolumn{2}{c}{\color{red}{\textbf{60}}}&\multicolumn{2}{c}{0}&\multicolumn{2}{c}{3}&\multicolumn{2}{c}{1}& \multicolumn{2}{c}{3}&\multicolumn{2}{c}{3}&\multicolumn{2}{c}{2}&\multicolumn{2}{c}{0}&\multicolumn{2}{c}{0}&\multicolumn{2}{c}{0}&\multicolumn{2}{c}{0}&\multicolumn{2}{c}{0}&\multicolumn{2}{c}{0}\\
			\hline
	\end{tabular}}
\end{table*}

\subsection{Evaluation Metrics}
To visually compare the forecasting performance of MPR-Net with other methods, we employ the mean square error (MSE) and mean absolute error (MAE) as evaluation metrics for long-term forecasting. And following the N-BEATS and TIMES-Net, we employ the symmetric mean absolute percentage error (SMAPE), mean absolute scaled error (MASE) and overall weighted average (OWA) as metrics for short-term forecasting. Above metrics can be calculated as:
\[{\rm{ SMAPE }} = \frac{{200}}{T}\sum\limits_{i = 1}^T {\frac{{\left| {{{\bf{X}}_i} - {{\widehat {\bf{X}}}_i}} \right|}}{{\left| {{{\bf{X}}_i}} \right| + \left| {{{\widehat {\bf{X}}}_i}} \right|}}}, {\rm{MAPE }} = \frac{{100}}{T}\sum\limits_{i = 1}^T {\frac{{\left| {{{\bf{X}}_i} - {{\widehat {\bf{X}}}_i}} \right|}}{{\left| {{{\bf{X}}_i}} \right|}}},\]
\[{\rm{ MASE }} = \frac{1}{T}\sum\limits_{i = 1}^T {\frac{{\left| {{{\bf{X}}_i} - {{\widehat {\bf{X}}}_i}} \right|}}{{\frac{1}{{T - m}}\sum\limits_{j = m + 1}^H {\left| {{{\bf{X}}_j} - {{\bf{X}}_{j - m}}} \right|} }}}, \]
\[{\rm{OWA}} = \frac{1}{2}\left[ {\frac{{{\rm{ SMAPE }}}}{{{\rm{SMAP}}{{\rm{E}}_{{\rm{N2 }}}}}} + \frac{{{\rm{ MASE }}}}{{{\rm{ MASE}}{{\rm{ }}_{{\rm{N}}2}}}}} \right],\]
where $m$ is the periodicity of the data. ${{{\widehat {\bf{X}}}}}$,${{\bf{X}}} \in {^{T \times D}}$ are the forecasting results and ground truth. $\bf{X}_i$ denotes the $i$-th future point.
\subsection{Comparison Methods}
In order to verify that MPR-Net achieves state-of-the-art performance, we compared it with current advanced and recognized models. These models include: (1) RNN-based models:LSTM (1997), LSSL (2022); (2) MLP-based models: LightTS (2022) and DLinear (2023); (3) Transformer-based models: LogTrans (2019), Reformer (2020), Informer (2021), Pyraformer (2021a), Autoformer (2021), FEDformer (2022), Non-stationary Transformer (2022a) and ETSformer (2022); (4) Decomposition-based models: TimesNet(2023), N-HiTS (2022) and N-BEATS (2019); (5) CNN-based Model: TCN (2019). In total, there are more than 16 baselines for comprehensive comparison.

\subsection{Experimental Setup}
MPR-Net is implemented through the Pytorch deep learning framework and trained on an Nvidia A40 GPU (48GB). To enhance the reproducibility of the implemented results, we fix random seeds. MPR-Net is built with 4 layers in long-term forecasting and 2 layers in short-term forecasting to accommodate inputs of different lengths. The model batch size is set to 32, training runs 10 epochs, and optimization is performed by Adam. Other parameters are set as in most comparison methods and are published in the open source code.

\begin{table*}[!ht]
	\renewcommand{\arraystretch}{1.4}
	\centering
	\caption{Complete results for the short-term forecasting task in M4 data set.}
	\resizebox{\textwidth}{30mm}{\setlength{\tabcolsep}{0.8mm}{
			\begin{tabular}{c c c c c c c c c c c c c c c c c c c}
				\hline
				\multicolumn{2}{c}{\multirow{2}{*}{Models}}&\textbf{MPR-Net}&TimesNet&N-HiTS&N-BEATS&ETS.&LightTS&DLinear&FED.&Stationary&Auto.&Pyra.&In.&LogTrans&Re.&LSTM&TCN&LSSL\\
				&&\textbf{(Ours)}&(2023)&(2022)&(2019)&(2022)&(2022)&(2023)&(2022)&(2022a)&(2021)&(2021a)&(2021)&(2019)&(2020)&(1997)&(2019)&(2022)\\
				\hline
				\multirow{3}{*}{\rotatebox{90}{Yearly}}&
				\multicolumn{1}{c|}{SMAPE}&\color{red}{\textbf{13.285}}&\color{blue}{\textbf{13.387}}&13.418&13.436&18.009&14.247&16.965&13.728&13.717&13.974&15.530&14.727&17.107&16.169&176.040&14.920&61.675\\
				&\multicolumn{1}{c|}{MASE}&\color{red}{\textbf{2.961}}&\color{blue}{\textbf{2.996}}&3.045&3.043&4.487&3.109&4.283&3.048&3.078&3.134&3.711&3.418&4.177&3.800&31.033&3.364&19.953\\
				&\multicolumn{1}{c|}{OWA}&\color{red}{\textbf{0.779}}&\color{blue}{\textbf{0.786}}&0.793&0.794&1.115&0.827&1.058&0.803&0.807&0.822&0.942&0.881&1.049&0.973&9.290&0.880&4.397\\
				\hline
				\multirow{3}{*}{\rotatebox{90}{Quarterly}}&
				\multicolumn{1}{c|}{SMAPE}&\color{red}{\textbf{10.011}}&\color{blue}{\textbf{10.100}}&10.202&10.124&13.376&11.364&12.145&10.792&10.958&11.338&15.449&11.360&13.207&13.313&172.808&11.122&65.999\\
				&\multicolumn{1}{c|}{MASE}&\color{red}{\textbf{1.167}}&1.182&1.194&\color{blue}{\textbf{1.169}}&1.906&1.328&1.520&1.283&1.325&1.365&2.350&1.401&1.827&1.775&19.753&1.360&17.662\\
				&\multicolumn{1}{c|}{OWA}&\color{red}{\textbf{0.88}}&0.890&0.899&\color{blue}{\textbf{0.886}}&1.302&1.000&1.106&0.958&0.981&1.012&1.558&1.027&1.266&1.252&15.049&1.001&9.436\\
				\hline
				\multirow{3}{*}{\rotatebox{90}{Monthly}}&
				\multicolumn{1}{c|}{SMAPE}&\color{red}{\textbf{12.668}}&\color{blue}{\textbf{12.670}}&12.791&12.677&14.588&14.014&13.514&14.260&13.917&13.958&17.642&14.062&16.149&20.128&143.237&15.626&64.664\\
				&\multicolumn{1}{c|}{MASE}&\color{red}{\textbf{0.933}}&\color{red}{\textbf{0.933}}&0.969&0.937&1.368&1.053&1.037&1.102&1.097&1.103&1.913&1.141&1.660&2.614&16.551&1.274&16.245\\
				&\multicolumn{1}{c|}{OWA}&\color{red}{\textbf{0.878}}&\color{red}{\textbf{0.878}}&0.899&\color{blue}{\textbf{0.880}}&1.149&0.981&0.956&1.012&0.998&1.002&1.511&1.024&1.340&1.927&12.747&1.141&9.879\\
				\hline
				\multirow{3}{*}{\rotatebox{90}{Others}}&
				\multicolumn{1}{c|}{SMAPE}&\color{red}{\textbf{4.714}}&\color{blue}{\textbf{4.891}}&5.061&4.925&7.267&15.880&6.709&4.954&6.302&5.485&24.786&24.460&23.236&32.491&186.282&7.186&121.844\\
				&\multicolumn{1}{c|}{MASE}&\color{blue}{\textbf{3.234}}&3.302&\color{red}{\textbf{3.216}}&3.391&5.240&11.434&4.953&3.264&4.064&3.865&18.581&20.960&16.288&33.355&119.294&4.677&91.650\\
				&\multicolumn{1}{c|}{OWA}&\color{red}{\textbf{1.006}}&\color{blue}{\textbf{1.035}}&1.040&1.053&1.591&3.474&1.487&1.036&1.304&1.187&5.538&5.879&5.013&8.679&38.411&1.494&27.273\\
				\hline
				\multirow{3}{*}{\rotatebox{90}{Average}}&\multicolumn{1}{c|}{SMAPE}&\color{red}{\textbf{11.774}}&\color{blue}{\textbf{11.829}}&11.927&11.851&14.718&13.525&13.639&12.840&12.780&12.909&16.987&14.086&16.018&18.200&160.031&13.961&67.156\\
				&\multicolumn{1}{c|}{MASE}&\color{red}{\textbf{1.571}}&\color{blue}{\textbf{1.585}}&1.613&1.599&2.408&2.111&2.095&1.701&1.756&1.771&3.265&2.718&3.010&4.223&25.788&1.945&21.208\\
				&\multicolumn{1}{c|}{OWA}&\color{red}{\textbf{0.845}}&\color{blue}{\textbf{0.851}}&0.861&0.855&1.172&1.051&1.051&0.918&0.930&0.939&1.480&1.230&1.378&1.775&12.642&1.023&8.021\\
				\hline
	\end{tabular}}}
\end{table*}

\subsection{Experimental Results for Long-Term Series Forecasting}
To compare the long-term forecasting performance of different models, we follow the setup commonly used in related works. Table 2 shows the results of the comparison between MPR-Net and other methods for long-term forecasting, with the best-performing metric values in red and the next best in blue.

The results clearly demonstrates that MPR-Net exhibits superior long-term forecasting performance compared to other methods. Out of the 72 forecastings made on 10 benchmark datasets, MPR-Net achieved the best results in 60 cases and achieved either the best or suboptimal results in 70 cases. In comparison to the current state-of-the-art model, TimesNet, MPR-Net not only outperforms it in all 72 forecastings but also significantly improves the forecasting performance on the ETT, Electricity, and particularly the traffic datasets. Notably, MPR-Net achieves an average forecasting MSE accuracy below 0.5 for the traffic dataset, which is a significant improvement over the existing comparison models where the forecasting MSE accuracy is generally higher than 0.6. The traffic dataset exhibits clear periodicity, and the experimental results suggest that MPR-Net can implicitly and effectively extract this periodicity information through adaptive convolutional pattern decomposition and the multi-scale hierarchical structure. By utilizing the extracted periodicity information for pattern reproduction, MPR-Net can generate more accurate forecastings. These findings verify that MPR-Net is capable of extracting and utilizing various valid patterns for forecasting purposes.

However, the advantage of our method is less pronounced when it comes to the Exchange dataset, and we can provide an explanation for this. In datasets where historical data exhibits fixed patterns, traders tend to exploit these patterns for their advantage, which ultimately disrupts the fixed pattern and impacts the market. Consequently, the forecasting accuracy of MPR-Net is affected in datasets with low predictability and less obvious pattern recurrence. In such cases, the fully-connected structure approach demonstrates a more powerful fitting ability by effectively modeling all information, resulting in better forecasting performance. Nonetheless, MPR-Net still achieves optimal or suboptimal forecasting accuracy in 6 out of 8 forecasting datasets.

To summarize, MPR-Net demonstrates excellent forecasting performance on time series data with predictable patterns. The leading forecasting performance across multiple real datasets further underscores the robustness of MPR-Net in long-term forecasting

\subsection{Experimental Results for Short-Term Series Forecasting}
To compare the short-term forecasting performance of the model, the forecasting length are shortened to half of the input length by referring to the settings of Timesnet and Nbeats. The forecasting lengths for all subset of M4 are: Yearly(6), Quarterly(8), Monthly(18), Weakly(13), Daily(14), Hourly(48). Table 3 reflects the short-term performance forecasting comparison between MPR-Net and other advanced methods.  

It can be verified from Table 3 that our proposed MPR-Net is equally good in short-term forecasting performance. Compared with other methods, MPR-Net achieves optimal results on 13/15 predictors for 6 sub-data and achieves optimal or suboptimal results on all predictors. Due to the short forecasting length of the M4 dataset, the fully connected structure can fit the forecasting target more easily at shorter forecasting lengths, thus improving the overall forecasting performance of most methods. Therefore, the forecasting performance gap between different methods is smaller and the forecasting performance improvement is more difficult. In this case, MPR-Net can still achieve the most advanced forecasting results further illustrating the effectiveness of the model construction. Combined with the state-of-the-art performance of MPR-Net in long-term forecasting, it can be demonstrated that through the design of multi-scale hierarchy, as well as adaptive pattern extraction at different scales and future forecasting based on pattern recurrence, the key patterns in the historical series can be truly extracted and effectively transformed into future forecasts, thus producing the best forecasting performance in both long- and short-term forecasting.

\begin{figure}
	\centering
	\includegraphics[width=85mm]{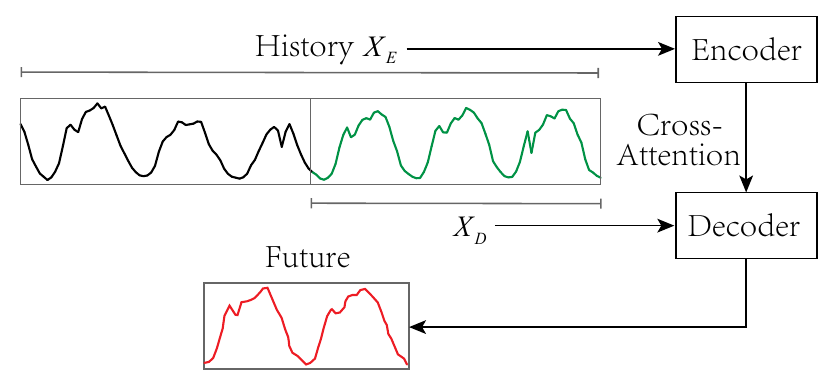}
	\caption{Common Transformer-based architectures for sime series forecasting.}
	\label{fig2:env}
\end{figure}

\subsection{Comparison of MPR-Net and Transformer-based Methods}
With the prominent performance of attention mechanism-based transformer methods in various fields, there has been a large amount of work applying them to time series forecasting. However, from the results of long- and short-term forecasting, the overall forecasting performance of the attention mechanism-based approach for the forecasting process is weaker than that of our proposed MPR-Net and MLP approaches. We further analyze this phenomenon. One significant feature of forecasting by attention mechanism compared with other methods is that the forecasting process is based on the decoder decoding of the encoder, as shown in Fig.4. These methods usually take the input complete history sequence as the input to the encoder and the second half of the history sequence as the input to the decoder. In this case the generation of future forecastings can only rely on the pattern features modeled by the second half of the history sequence extracted, which leads to a decrease in decoding power when the key patterns of future reproduction cannot be fully contained by the second half of the history sequence, resulting in less accurate forecastings.

In contrast to them, MPR-Net benefits from a multiscale hierarchy in which longer-length patterns at lower levels can be captured by shorter-length patterns at higher levels. Therefore, in the pattern matching corresponding to the above decoding process, longer pattern matching at lower levels can be accomplished by shorter pattern matching at higher levels while reducing the computational expenditure. When the number of layers is large enough, the higher-level patterns can capture the lower-level patterns in the whole historical sequence, solving the above-mentioned problem. Meanwhile, MPR-Net does not use the structure of point-by-point connection of attention mechanism, which preserves the temporal information and improves the utilization of information in the historical sequence, explaining its better forecasting performance from another aspect.

\begin{figure}
	\begin{minipage}[t]{0.25\textwidth}
		\centering
		\includegraphics[width=1.8in]{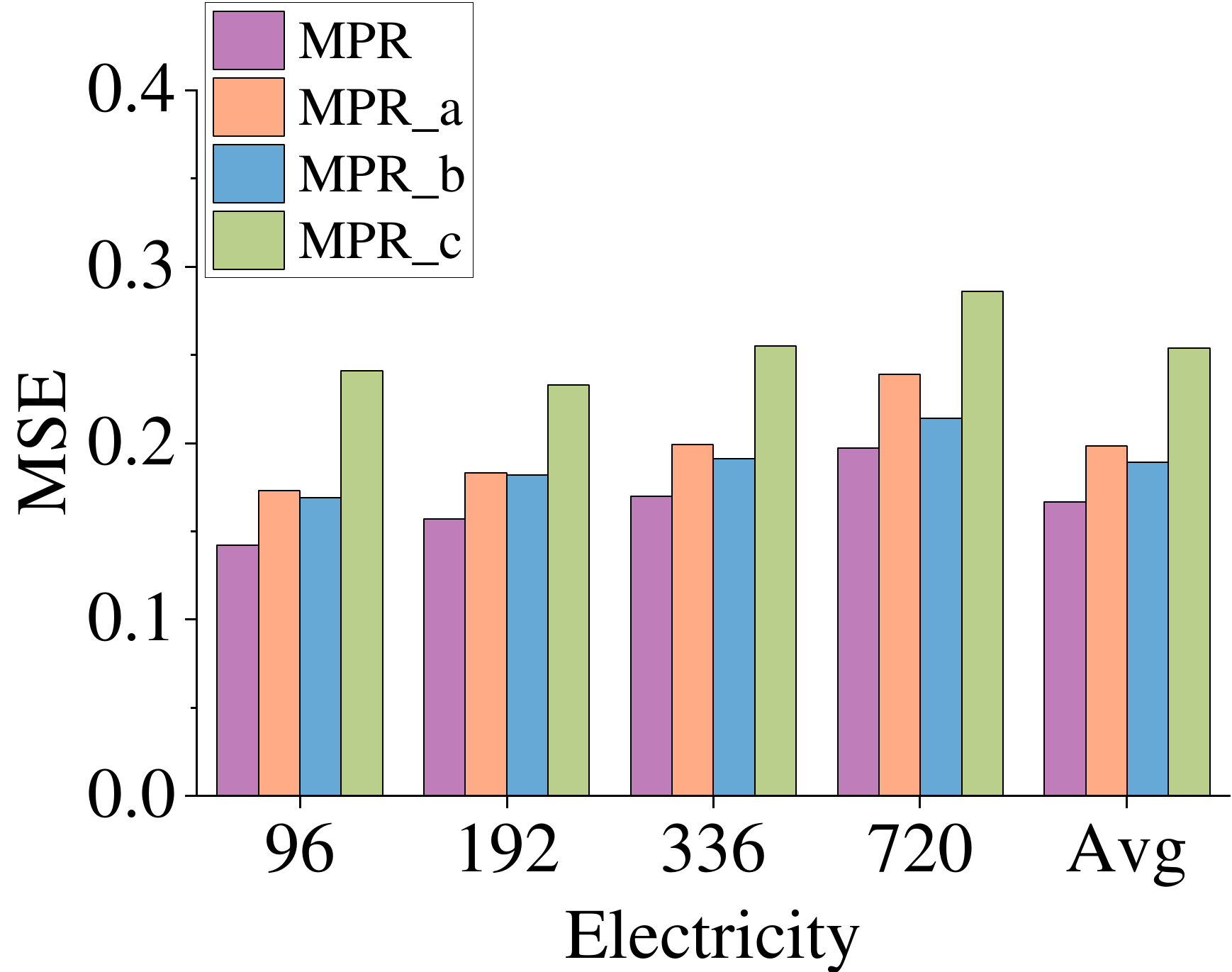}
		\label{fig:side:a}
	\end{minipage}%
	\begin{minipage}[t]{0.25\textwidth}
		\centering
		\includegraphics[width=1.8in]{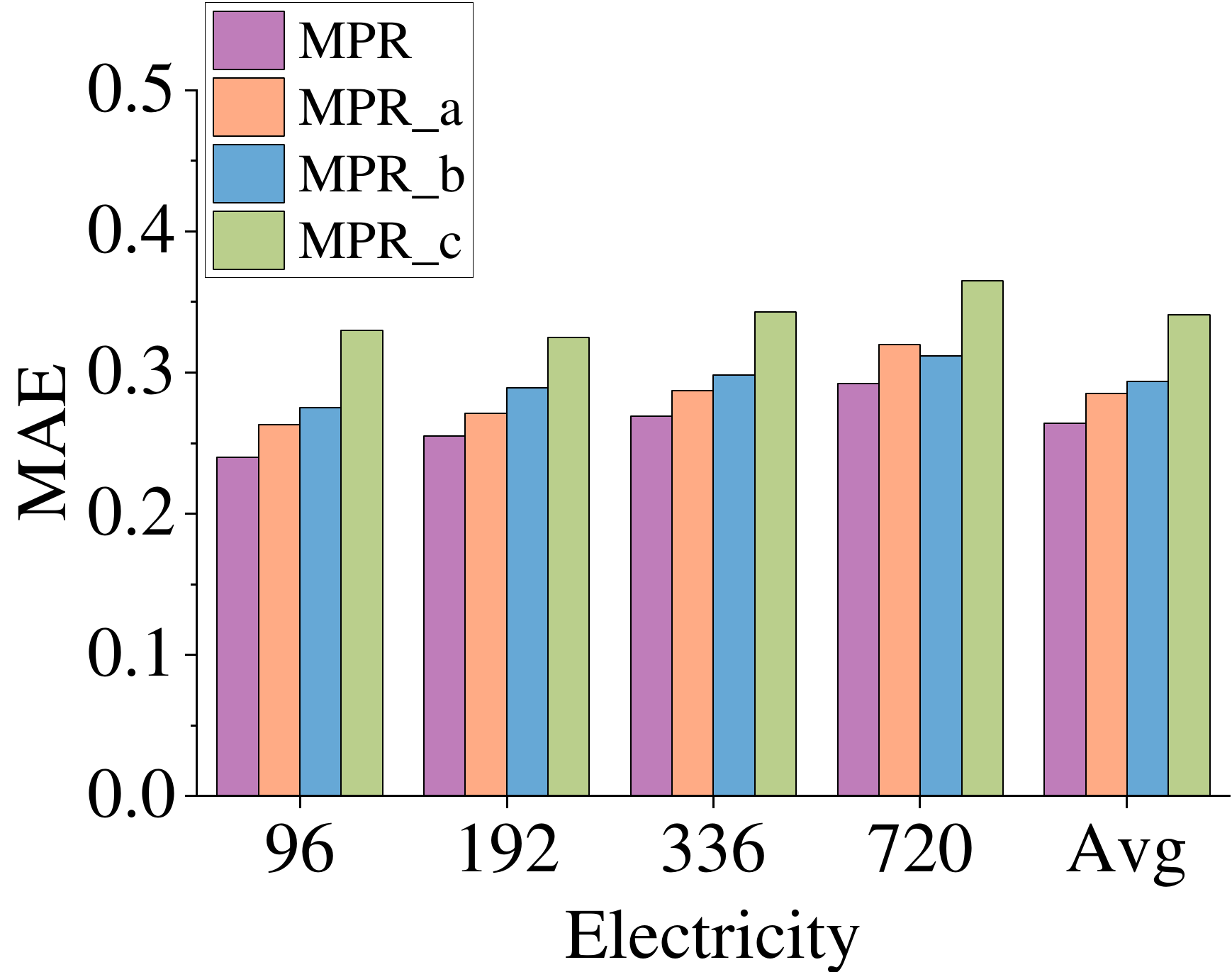}
		\label{fig:side:b}
	\end{minipage}
	\begin{minipage}[t]{0.25\textwidth}
		\centering
		\includegraphics[width=1.8in]{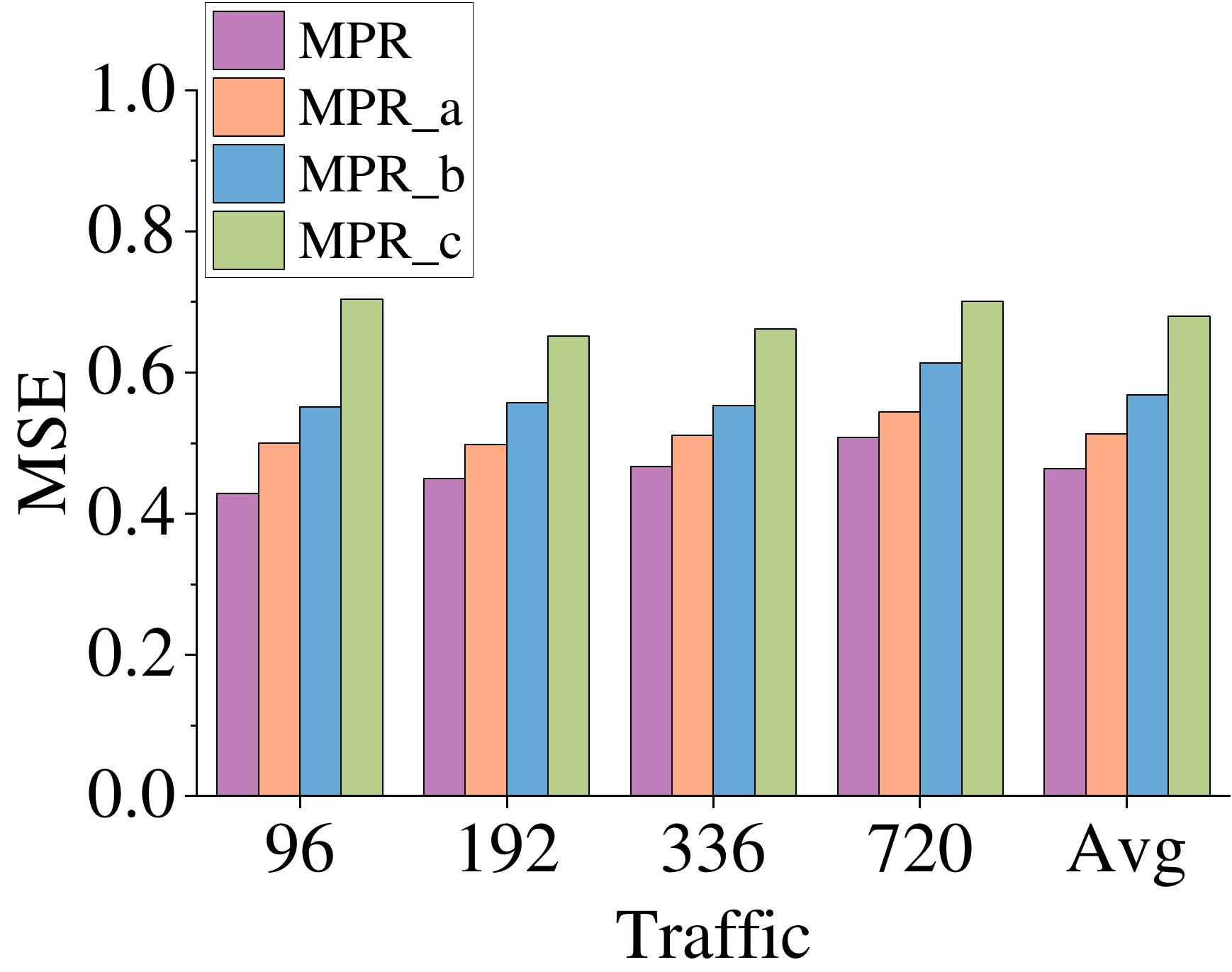}
		\label{fig:side:a}
	\end{minipage}%
	\begin{minipage}[t]{0.25\textwidth}
		\centering
		\includegraphics[width=1.8in]{ab_tra_mse_bigbig.pdf}
		\label{fig:side:b}
	\end{minipage}
	\caption{Ablation experiment performed on Electircity and Traffic data sets.}
\end{figure}

\subsection{Ablation Experiments}
The proposed MPR-Net model relies mainly on the fusion and interaction of patterns extracted from multiple convolutional kernels across variables, and pattern extension forecasting based on pattern recurrence. To verify the impact of these two important components on the model performance, three sub-models are constructed:

MPR-Net\_a performs independent convolution and transposition convolution operations on each channel, neglecting the fusion and interaction of patterns across variables. MPR-Net\_b relies on a simple fully connected layer without pattern extension forecasting. MPR-Net\_c combines the approaches of MPR-Net\_a and MPR-Net\_b.

The comparative analysis depicted in Figure 5 reveals the superior forecasting performance of MPR-Net over the three sub-models. This finding emphasizes the crucial role played by two key components of MPR-Net: the fusion and interaction of patterns derived from convolutional kernels of multiple variables, and the pattern expansion forecasting based on pattern reproduction. Comparing the two datasets, MPR-Net\_a outperforms MPR-Net\_b in forecasting traffic data, while the reverse is observed for Electircity data. This discrepancy suggests that the number of valid patterns, variable relationships, and the degree of pattern reproduction in the series may differ across datasets. However, effective modeling of this information can sufficiently enhance forecasting performance.

Additionally, MPR-Net\_c, which extracts patterns through channel-independent convolution and utilizes a fully connected layer for future forecasting, exhibits the poorest performance at different forecasting lengths for both datasets. This highlights the existence of multiple patterns within time series data and their potential reproduction in the future. The information pertaining to these patterns can be strengthened through fusion and interaction between variables. In comparison to MPR-Net\_a, incorporating a priori knowledge based on pattern recurrence significantly improves forecasting accuracy by extending the patterns. Importantly, MPR-Net's forecasting process achieves lower time complexity and enhanced interpretability compared to using solely a fully connected layer.

\begin{table}[]
	\centering
	\caption{Impact of model layers on long-term forecasting performance.}
	\resizebox{90mm}{22mm}{
		\renewcommand{\arraystretch}{1.5}
		\begin{tabular}{|c|c|cc|cc|cc|cc|}
			\hline
			\multirow{2}{*}{Dataset} &
			\multirow{2}{*}{Length} &
			\multicolumn{2}{c|}{layer2} &
			\multicolumn{2}{c|}{layer3} &
			\multicolumn{2}{c|}{layer4} &
			\multicolumn{2}{c|}{layer5} \\ \cline{3-10} 
			&
			&
			\multicolumn{1}{c|}{MSE} &
			MAE &
			\multicolumn{1}{c|}{MSE} &
			MAE &
			\multicolumn{1}{c|}{MSE} &
			MAE &
			\multicolumn{1}{c|}{MSE} &
			MAE \\ \hline
			\multirow{4}{*}{ETTh1} &
			96 &
			\multicolumn{1}{c|}{0.385} &
			0.398 &
			\multicolumn{1}{c|}{0.383} &
			0.397 &
			\multicolumn{1}{c|}{0.383} &
			0.398 &
			\multicolumn{1}{c|}{0.382} &
			0.397 \\ \cline{2-10} 
			&
			192 &
			\multicolumn{1}{c|}{0.44} &
			0.428 &
			\multicolumn{1}{c|}{0.438} &
			0.427 &
			\multicolumn{1}{c|}{0.435} &
			0.427 &
			\multicolumn{1}{c|}{0.434} &
			0.426 \\ \cline{2-10} 
			&
			336 &
			\multicolumn{1}{c|}{0.485} &
			0.452 &
			\multicolumn{1}{c|}{0.483} &
			0.45 &
			\multicolumn{1}{c|}{0.475} &
			0.446 &
			\multicolumn{1}{c|}{0.475} &
			0.446 \\ \cline{2-10} 
			&
			720 &
			\multicolumn{1}{c|}{0.508} &
			0.484 &
			\multicolumn{1}{c|}{0.48} &
			0.466 &
			\multicolumn{1}{c|}{0.49} &
			0.471 &
			\multicolumn{1}{c|}{0.495} &
			0.478 \\ \hline
			\multirow{4}{*}{ETTm2} &
			96 &
			\multicolumn{1}{c|}{0.177} &
			0.259 &
			\multicolumn{1}{c|}{0.176} &
			0.258 &
			\multicolumn{1}{c|}{0.172} &
			0.255 &
			\multicolumn{1}{c|}{0.176} &
			0.256 \\ \cline{2-10} 
			&
			192 &
			\multicolumn{1}{c|}{0.242} &
			0.3 &
			\multicolumn{1}{c|}{0.241} &
			0.3 &
			\multicolumn{1}{c|}{0.238} &
			0.298 &
			\multicolumn{1}{c|}{0.246} &
			0.304 \\ \cline{2-10} 
			&
			336 &
			\multicolumn{1}{c|}{0.304} &
			0.34 &
			\multicolumn{1}{c|}{0.299} &
			0.338 &
			\multicolumn{1}{c|}{0.297} &
			0.337 &
			\multicolumn{1}{c|}{0.31} &
			0.344 \\ \cline{2-10} 
			&
			720 &
			\multicolumn{1}{c|}{0.411} &
			0.403 &
			\multicolumn{1}{c|}{0.412} &
			0.406 &
			\multicolumn{1}{c|}{0.399} &
			0.397 &
			\multicolumn{1}{c|}{0.419} &
			0.406 \\ \hline
	\end{tabular}}
\end{table}

\begin{table}[]
	\centering
	\caption{Impact of model layers on short-term forecasting performance.}
	\resizebox{80mm}{16mm}{
		\renewcommand{\arraystretch}{1.5}
		\begin{tabular}{|c|c|c|c|c|c|}
			\hline
			Dataset                       & Metric & layer1 & layer2 & layer3 & layer4 \\ \hline
			\multirow{3}{*}{M4-Yearly}    & SMAPE  & 13.448 & 13.285 & 13.435 & 13.419 \\ \cline{2-6} 
			& MASE   & 3.034  & 2.961  & 3.026  & 3.046  \\ \cline{2-6} 
			& OWA    & 0.793  & 0.779  & 0.792  & 0.794  \\ \hline
			\multirow{3}{*}{M4-Quarterly} & SMAPE  & 10.085 & 10.011 & 10.016 & 10.009 \\ \cline{2-6} 
			& MASE   & 1.185  & 1.167  & 1.169  & 1.167  \\ \cline{2-6} 
			& OWA    & 0.89   & 0.88   & 0.881  & 0.88   \\ \hline
	\end{tabular}}
\end{table}

\subsection{Hyperparameter Analysis}
Since MPR-Net relies on multi-scale pattern decomposition, we further argue the effect of the defferent layer structure on the forecasting performance.

Table 4 and Table 5 present the impact of the number of layers on the forecasting performance of MPR-Net. In the case of long-term forecasting on the ETTh1 and ETTm2 datasets, the forecasting performance of patterns extracted from layers 2 and 3 are comparable, suggesting that patterns at the third layer scale have limited influence on forecasting and are not the primary drivers of future forecasting. However, with MPR-Net's 4 layers configuration, a significant improvement in forecasting accuracy is observed, indicating the extraction of more crucial patterns at this scale that guide future sequence changes. On the other hand, when MPR-Net is extended to 5 layers, a slight decline in forecasting accuracy is observed. This decline may be attributed to the model having already captured sufficient valid information for future prediction within the first four layers, while the additional information generated at the fifth layer scale introduces redundancy that affects the forecasting results. Hence, for long-term forecasting with a given input sequence length, a 4-layer MPR-Net configuration is chosen as it yields optimal forecasting performance.

For short-term prediction, due to the short input data, MPR-Net can capture patterns at different scales in the data with fewer layers and utilize them for future forecasting. In Table 5, on the M4-yearly and M4-Quarterly datasets, the forecasting performance of MPR-Net with 1 layer is inferior to that of 2 layers. This suggests that patterns extracted at a small scale in a single layer are insufficient for accurate predictions, leading to underfitting. However, increasing the number of layers beyond 2 does not significantly improve the forecasting performance and can even result in a decline, as observed in M4-yearly. This indicates that for short-term forecasting, the second layer scale is capable of capturing the key patterns from the input, and including too many layers may harm the model performance. Therefore, a 2-layer MPR-Net is constructed for short-term forecasting.

\begin{figure}
	\begin{minipage}[t]{0.25\textwidth}
		\centering
		\includegraphics[width=1.6in]{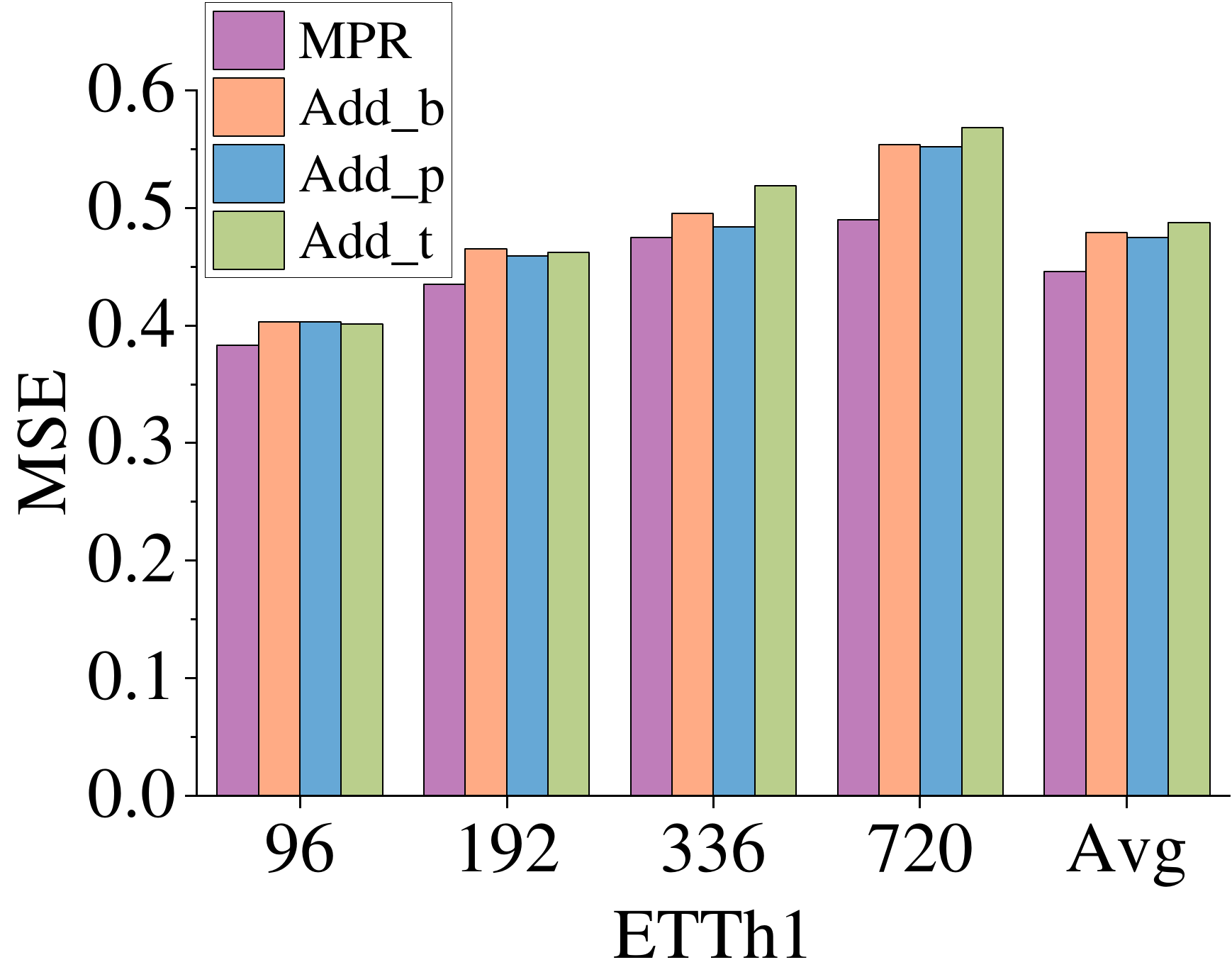}
		\label{fig:side:a}
	\end{minipage}%
	\begin{minipage}[t]{0.25\textwidth}
		\centering
		\includegraphics[width=1.6in]{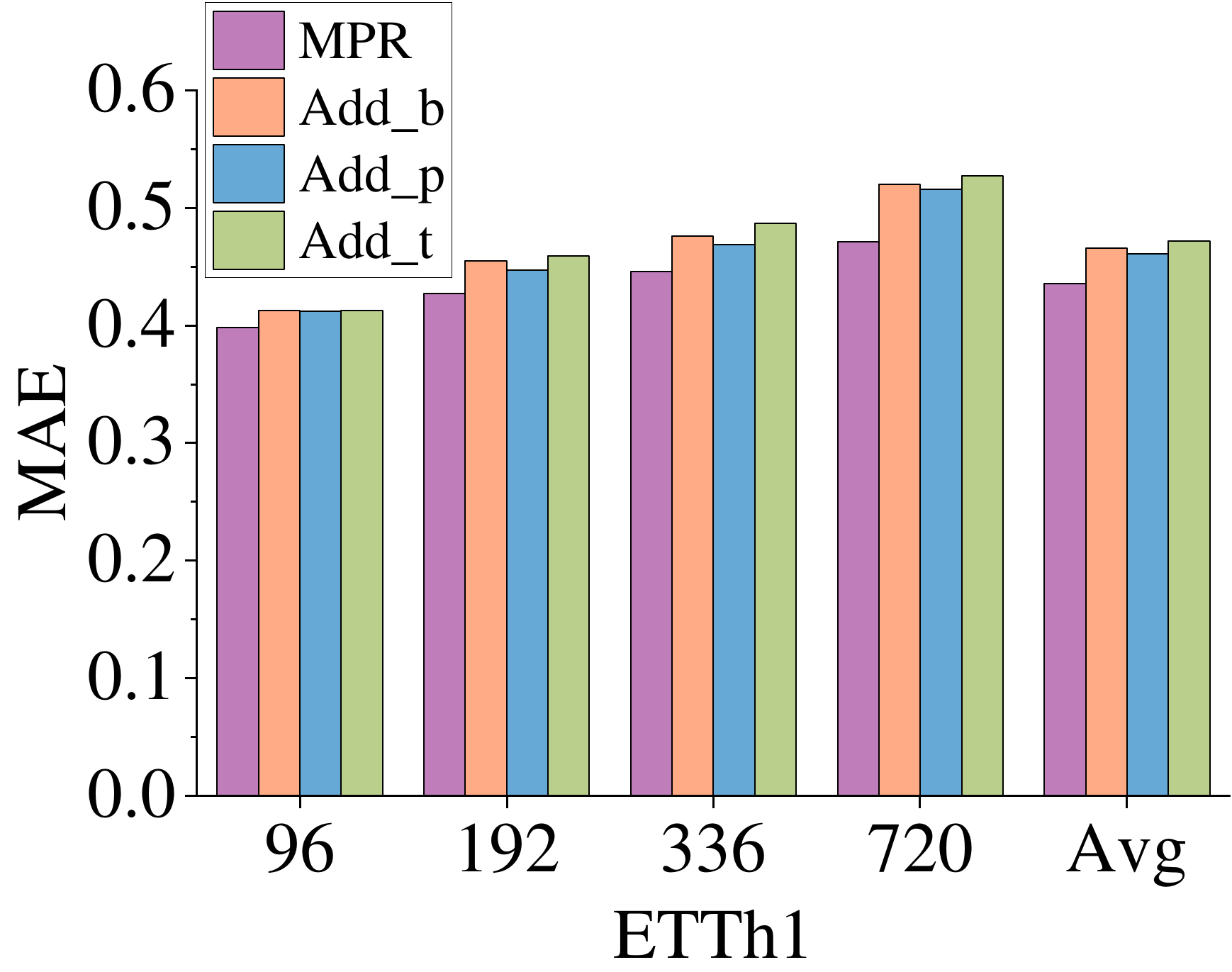}
		\label{fig:side:b}
	\end{minipage}
	\begin{minipage}[t]{0.25\textwidth}
		\centering
		\includegraphics[width=1.6in]{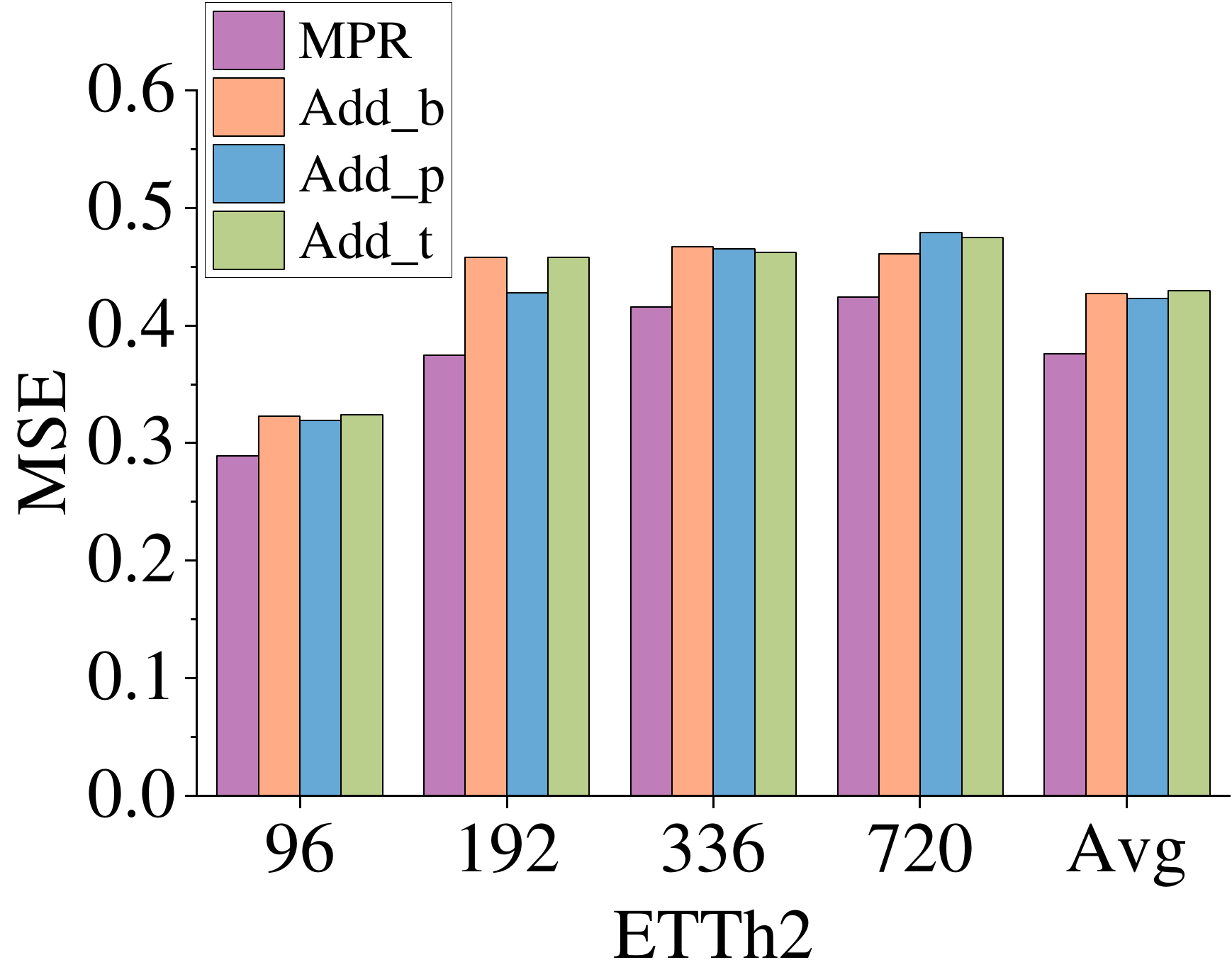}
		\label{fig:side:a}
	\end{minipage}%
	\begin{minipage}[t]{0.25\textwidth}
		\centering
		\includegraphics[width=1.6in]{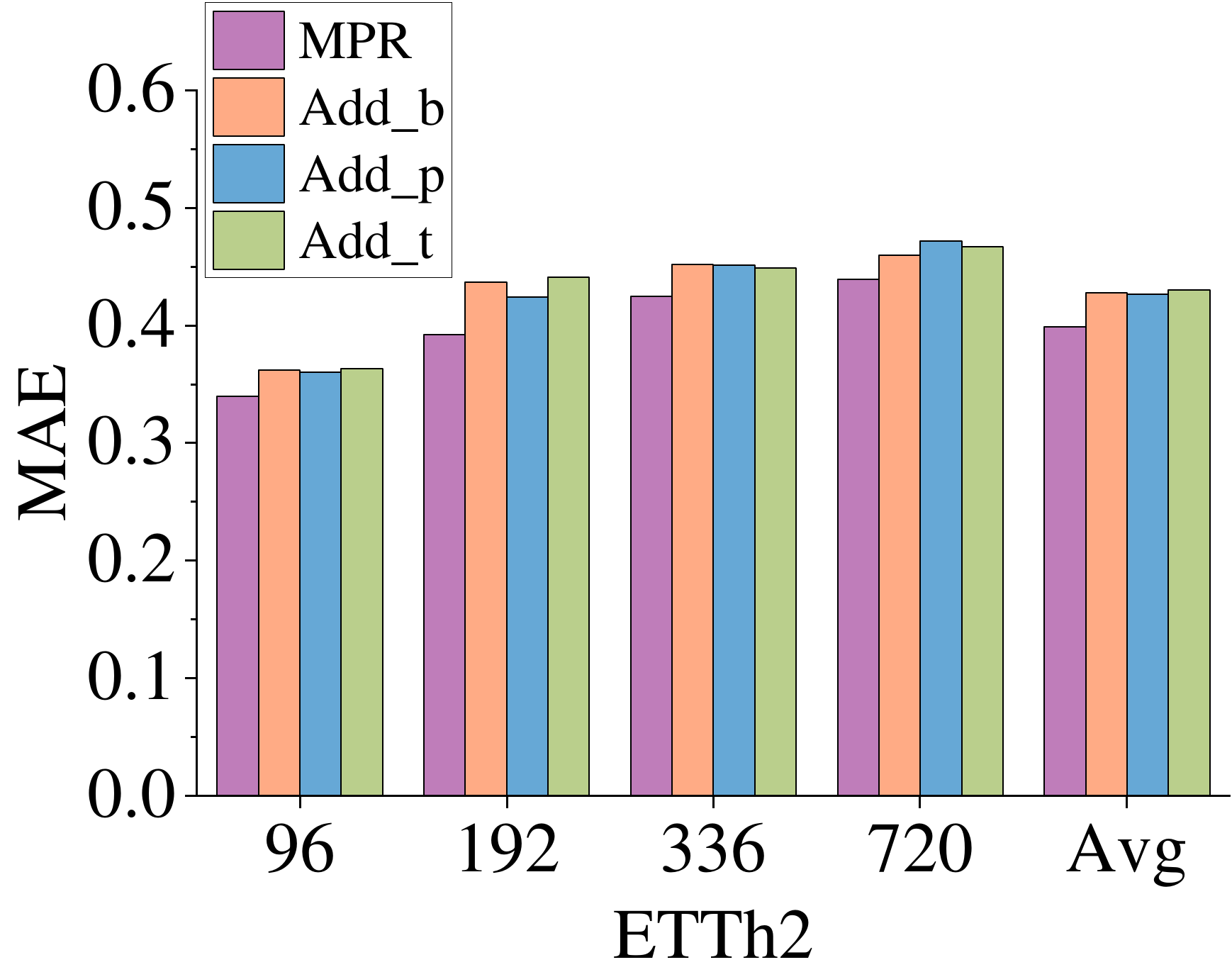}
		\label{fig:side:b}
	\end{minipage}
	\caption{Effects of temporal dependence encoding on forecasting performance}
\end{figure}

\subsection{Impact of Adding Temporal Information}

To address the issue of temporal information corruption in existing MLP and transformer-based methods, many approaches incorporate location encoding and temporal encoding information. In this study, we explore the impact of incorporating temporal information on the forecasting performance of MPR-Net.
We conducted experiments on the ETTh1 and ETTh2 datasets using three variants of MPR-Net: Add\_p, Add\_t, and Add\_b. Add\_p includes location encoding, Add\_t includes temporal encoding, and Add\_b includes both location and temporal encoding in the original model inputs.

The results demonstrate in Fig.6 that Add\_p performs slightly better than Add\_t, indicating that introducing temporal information through location encoding is more effective than temporal encoding alone for both datasets. Add\_b performs between the other two variants, suggesting that introducing both location and temporal encoding information does not further enhance the forecasting performance compared to using location encoding alone. These findings validate the advantages of MPR-Net's design, where pattern extraction and reconstruction through convolution and deconvolution operations, along with pattern extension forecasting based on prior knowledge of pattern recurrence, effectively preserve and utilize the temporal information in the original data.
The inclusion of additional temporal information through location and temporal encoding does not contribute to improved forecasting performance in MPR-Net. Instead, it introduces perturbations and reduces the accuracy of the forecasts. Thus, the unique network structure of MPR-Net enables the development of more accurate forecasts while preserving the temporal information of the historical series, distinguishing it from other existing methods.

\begin{figure}
	\begin{minipage}[t]{0.25\textwidth}
		\centering
		\includegraphics[width=1.6in]{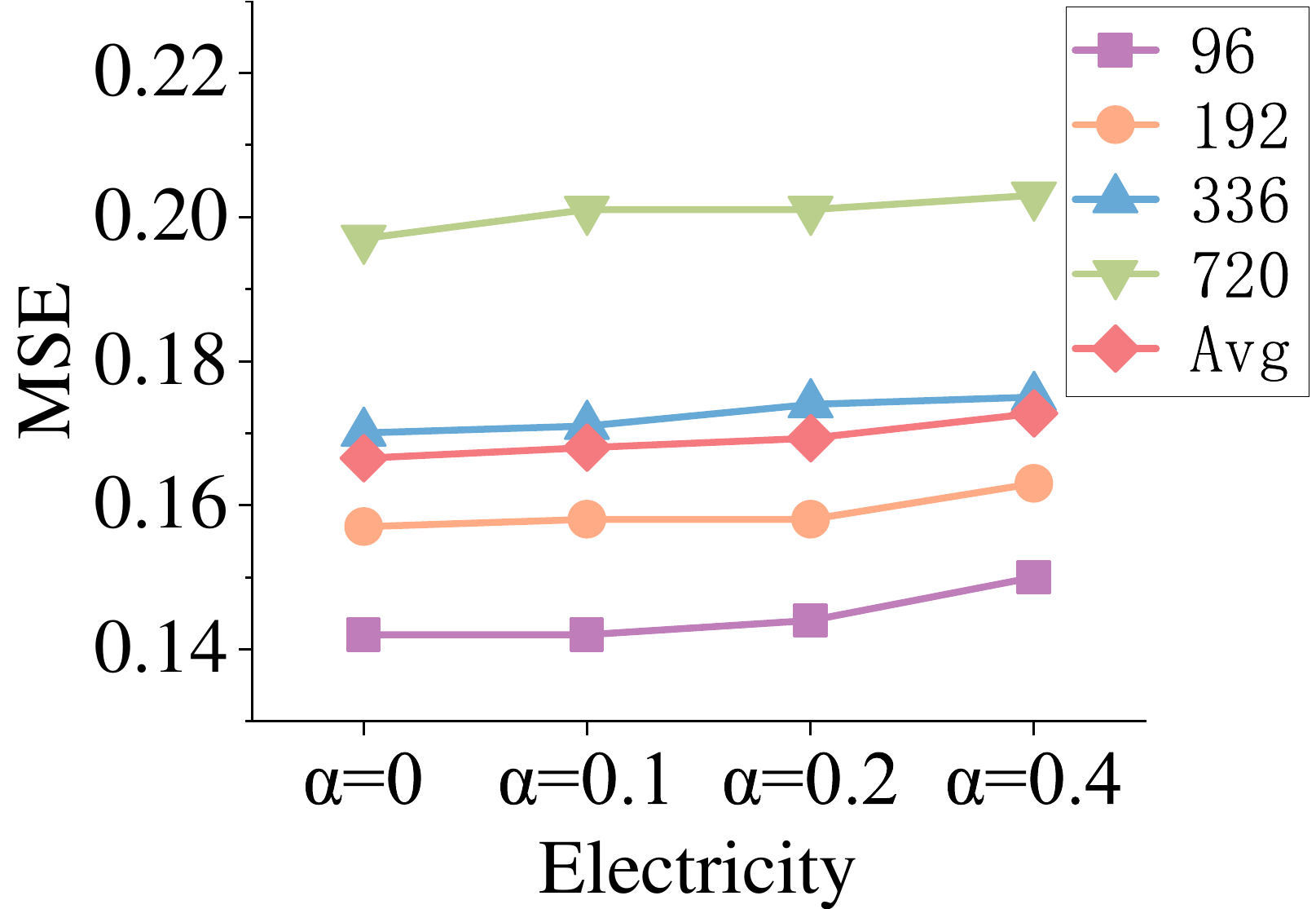}
		\label{fig:side:a}
	\end{minipage}%
	\begin{minipage}[t]{0.25\textwidth}
		\centering
		\includegraphics[width=1.6in]{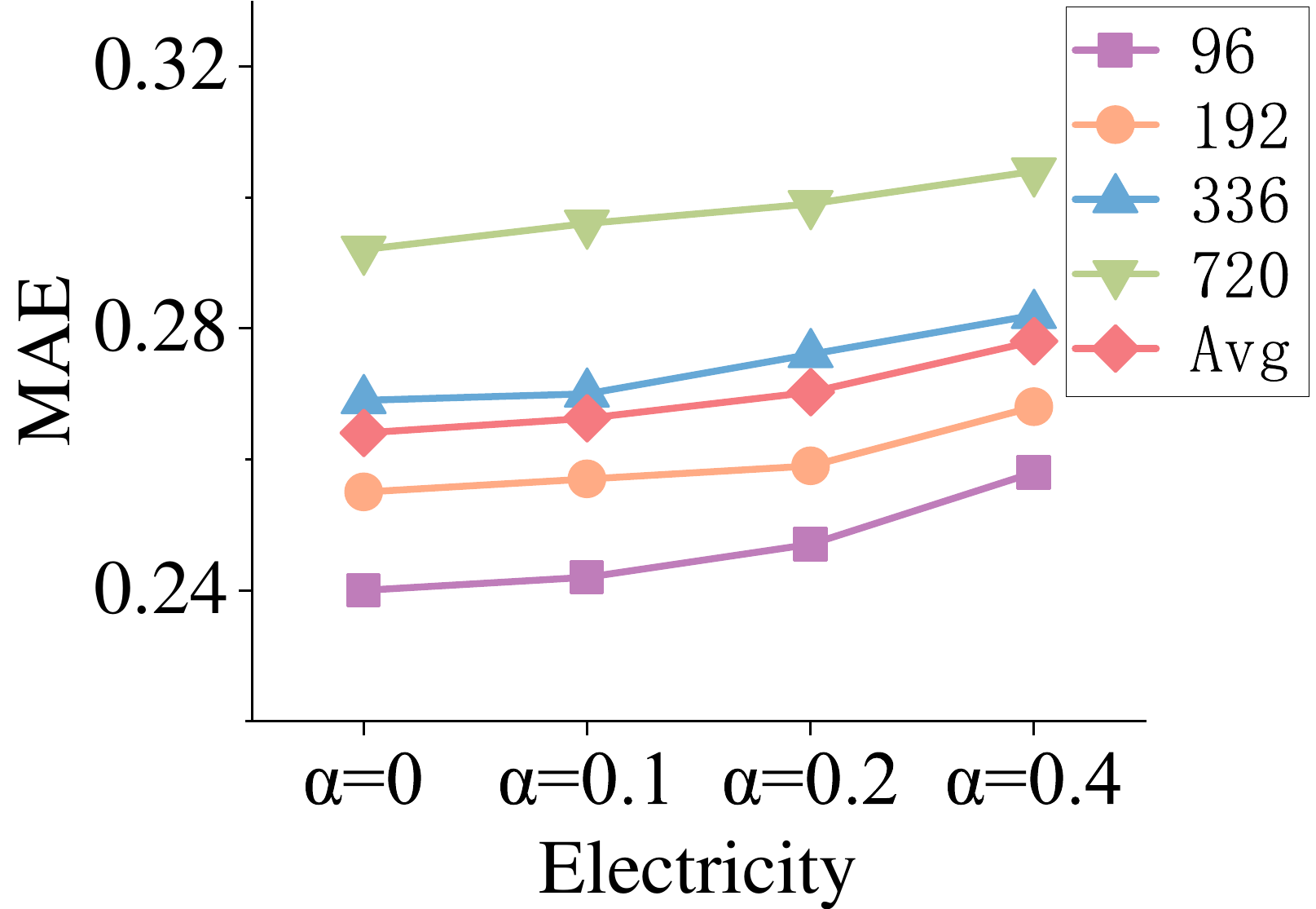}
		\label{fig:side:b}
	\end{minipage}
	\begin{minipage}[t]{0.25\textwidth}
		\centering
		\includegraphics[width=1.6in]{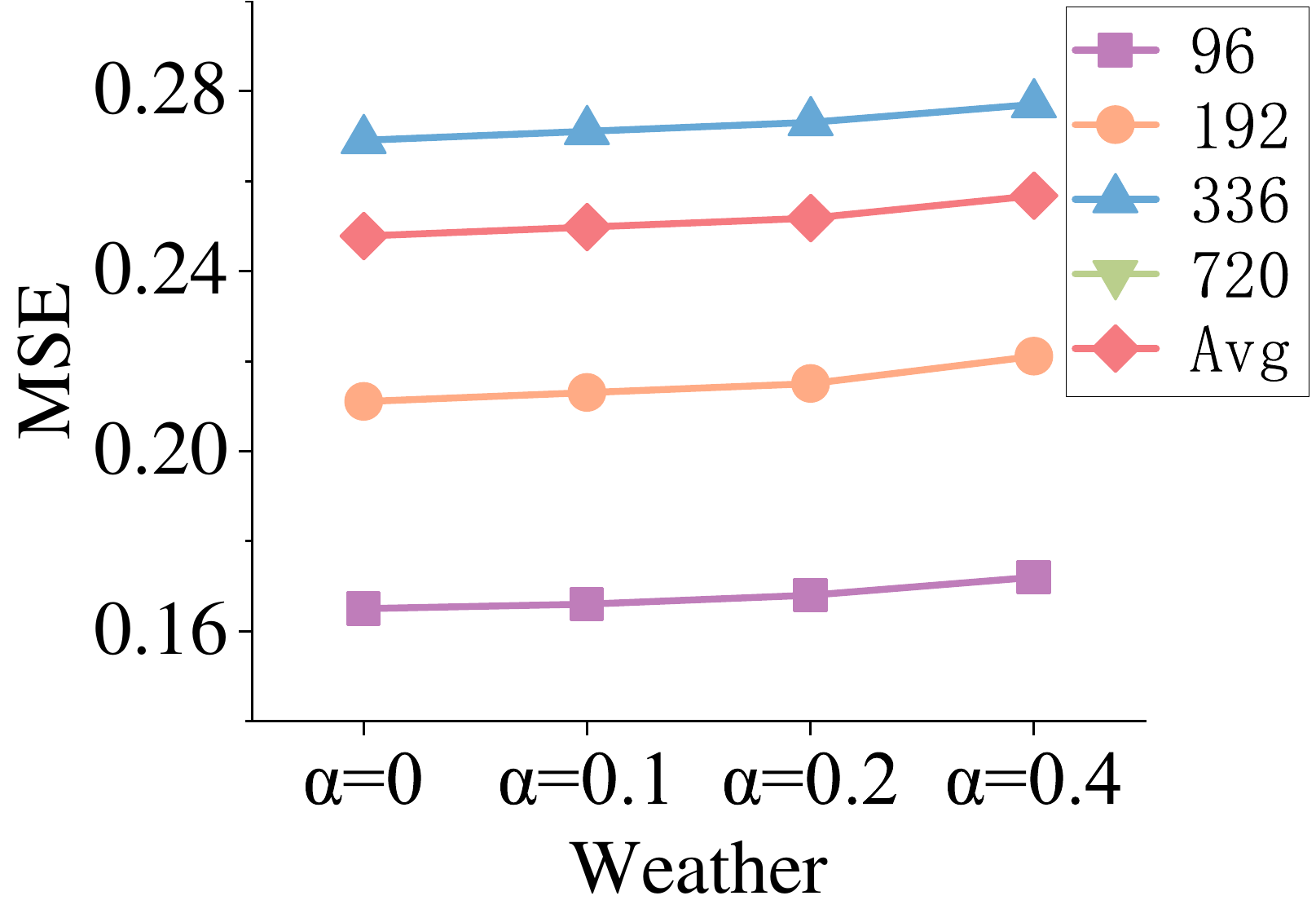}
		\label{fig:side:a}
	\end{minipage}%
	\begin{minipage}[t]{0.25\textwidth}
		\centering
		\includegraphics[width=1.6in]{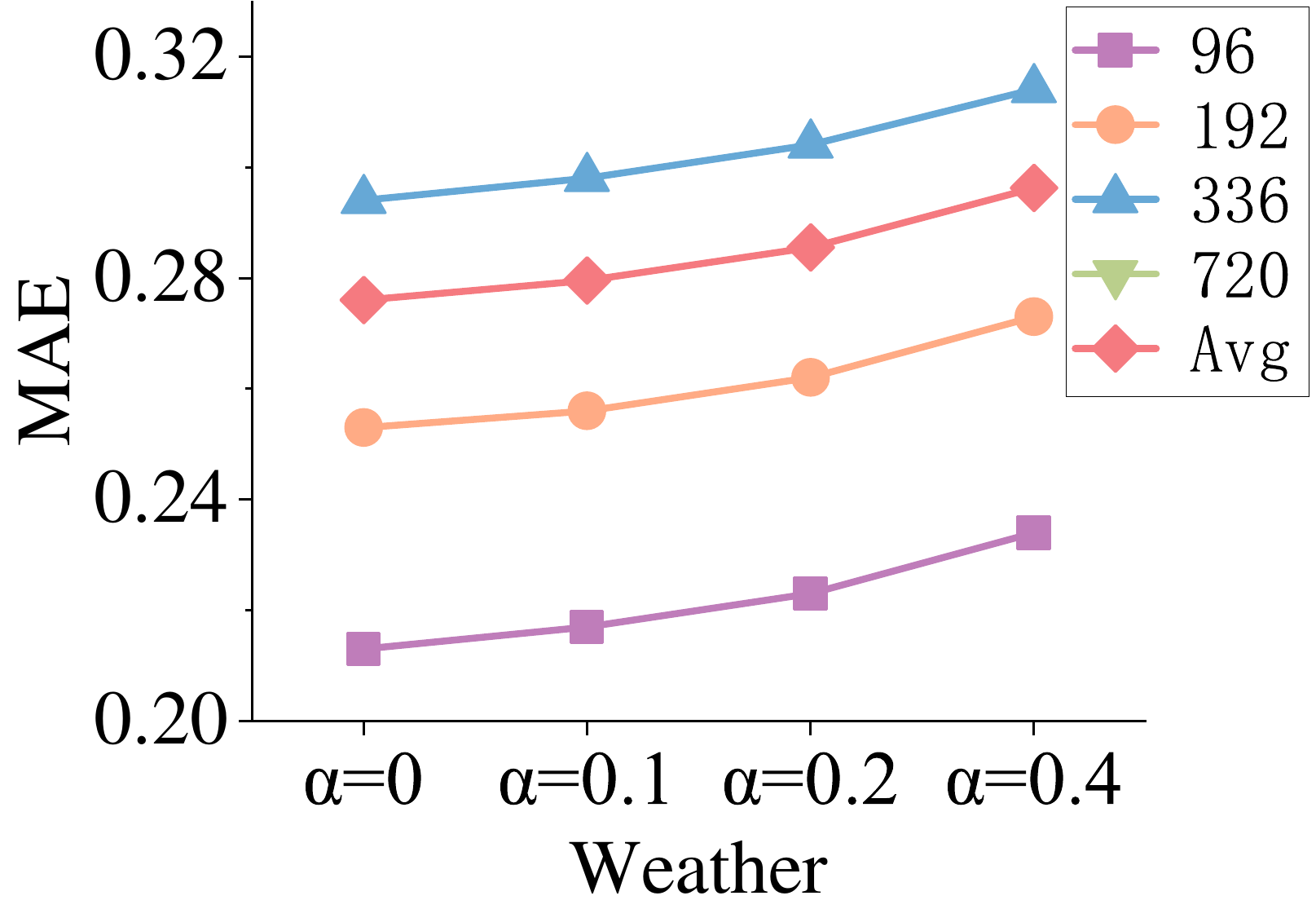}
		\label{fig:side:b}
	\end{minipage}
	\caption{Effects of different noise scales on forecasting performance}
\end{figure}

\subsection{Impact of Adding Noise Information}
In the previous section, when adding sequential information to MPR-Net by means of position encoding and time encoding, it is necessary to encode the input historical sequence values at the same time. In this way, the small gap between the values of the three encodings makes the original data information strong perturbed, resulting in a severe impairment of the forecasting performance. Therefore, to verify the performance of MPR-Net under appropriate noise, we add different scales of noise to the raw input data to study its effect on the forecasting performance. We add 0.1, 0.2, and 0.4 Gaussian noise to the input data according to its variance, which is expressed by:
\[{I_{{\rm{in}}}} = I + S \times \alpha  \times \mu ,\]
where $I$ means the input data, $S$ means the standard deviation of $I$, $\alpha$ is the noise level coefficient, and $\mu  \sim N\left( {0,1} \right)$ is a random variable with normal distribution.

Through Fig.7 we can find that the MPR-Net model is robust to small-scale noise on different data sets. The forecasting performance of MPR-Net decreases sluggishly at different noise levels: 0.1, 0.2, and 0.4. Adding 0.1 noise to the data has almost no effect on the forecasting performance. When the noise is expanded to 0.4, the forecasting performance is still acceptable. Based on the experimental results, we believe that the adaptive pattern extraction and series reconstruction block built by MPR-Net can effectively extract and reconstruct the key patterns in history and future, excluding the noisy information, thus reducing its interference with the data. In this way, we further validate the robustness of MPR-Net with its insensitive performance to noise.

\begin{table}[]
	\centering
	\caption{The time complexity comparison of MPR-Net and other Models.}
	\resizebox{80mm}{16mm}{
		\renewcommand{\arraystretch}{1.5}
	\begin{tabular}{|c|c|c|c|c|c|}
		\hline
		\multirow{2}{*}{Methods} & Training & Testing & \multirow{2}{*}{Methods} & Training   & Testing \\ \cline{2-3} \cline{5-6} 
		& Time     & Steps   &                          & Time       & Steps   \\ \hline
		MPR-Net                  & $O(n)$     & 1       & FEDformer                & $O(n)$       & 1       \\ \hline
		TimesNet                 & $O({n^2})$         & 1        & Autoformer               & $O(n log n)$ & 1       \\ \hline
		DLinear                  &  $O({n^2})$        &   1      & Informer                 & $O(n log n)$ & 1       \\ \hline
		 N-HiTS                  &  $O({n^2})$        &  1       & LogTrans                 & $O(n log n)$ & 1       \\ \hline
		N-BEATS                  & $O({n^2})$     &   1      & Reformer                 & $O(n log n)$ & $n$       \\ \hline
		TCN                      & $O({n^2})$     & 1       & LSTM                     & $O(n)$       & $n$       \\ \hline
	\end{tabular}}
\end{table}
\subsection{Time Complexity Analysis and Comparison}
We further analyze the time complexity of the model for the overall process of MPR-Net, which consists of three main blocks: HFE, PEF, and FFR. In the HFE and FFR blocks, the convolution and transposed convolution processes in the temporal dimension and the correlation attention process in the channel dimension are mainly included. Referring to most work, which treat the length of the sequence as a variable $n$ and the number of multivariate as a constant $d$. Then the time complexity of the convolution and transposed convolution process is $d \times n$, approximated by $O(n)$, and the time complexity of the correlation attention is ${d^2} \times n$, approximated by $O(n)$. In the PEF blocks, compared to the simple splicing in pattern extensions, the main computational expense is in pattern correlation matching. Pattern correlation matching requires matrix computation of $Q$ and $K$. If the length of $Q$ is small enough, the time complexity of its attention computation is $l \times (n - l)$, approximated by $O(n)$, where $l$ is the length of q and $(n - l)$ is the length of $K$. However, too short the length of $Q$ will result in insufficient pattern recognition, i.e., the length of the pattern is larger than the length of $Q$. Benefit from the design of the multi-scale hierarchy of MPR-Net, in the higher-level PEF blocks, although the length of $Q$ is short, it has a large perceptual field, so that patterns with large length in the lower level can be extracted, avoiding the above-mentioned possible problems. In summary, the time complexity of the proposed MPR-Net is approximately $O(n)$, which is ahead of most current work in time series forecasting.

Table 6 summarizes the comparison of time complexity in training and inference steps in testing. It can be seen that the proposed MPR-Net achieves the best overall complexity among the MLP and Transformer based forecasting models.

\subsection{Time Series Forecasting Results Presentation}
To show the forecasting performance of MPR-Net more visually as shown in Fig.8, we visualized its forecasting results on six different datasets. Observing these datasets we can find that datasets such as traffic and electricity show more obvious pattern information and the distribution of data is more regular. On the contrary, the ETTm1 dataset shows obvious noise with distance fluctuations in a small range, while the ILL dataset shows no obvious patterns, and they increase the difficulty of forecasting.

The forecasting results demonstrate the excellent forecasting ability of MPR-Net on different data. For data sets with low noise and obvious patterns, MPR-Net can accurately capture the periodic and trend information in the data set by pattern extension on the one hand. On the other hand, MPR-Net can accurately extract and use the patterns for forecasting of some abrupt changes in traffic and power datasets, thanks to the adaptive pattern extraction. For the noisy ETTm1 and ILL datasets, where the patterns are not obvious, MPR-Net can also extract the main patterns, i.e., trend information, from these hard-to-predict data that will be reproduced in the future. These forecasting results further validate the excellent performance of MPR-Net in time series forecasting, and its generalization in forecasting on different datasets is demonstrated.

\begin{figure}
	\begin{minipage}[t]{0.26\textwidth}
		\centering
		\includegraphics[width=1.9in]{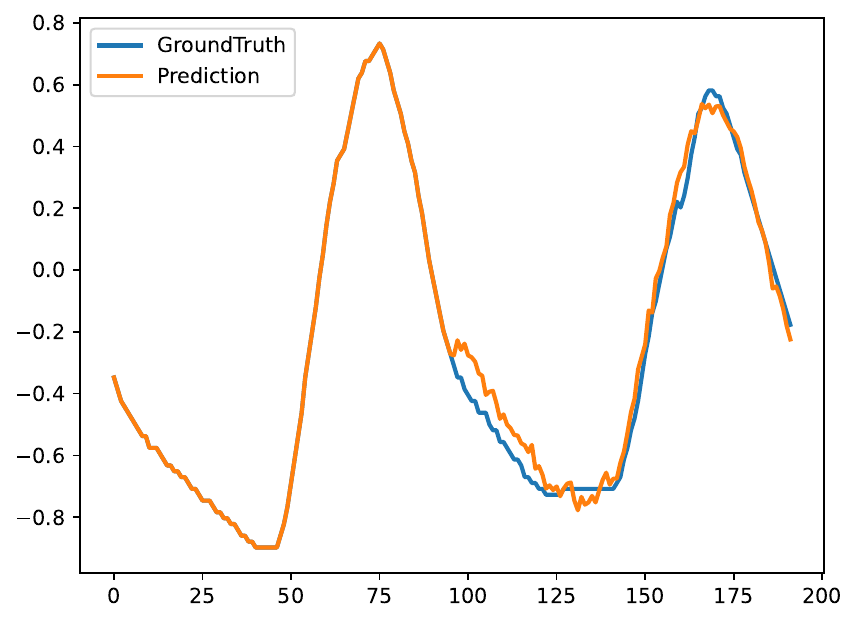}
		\caption*{ETTm2}
	\end{minipage}%
	\begin{minipage}[t]{0.26\textwidth}
		\centering
		\includegraphics[width=1.9in]{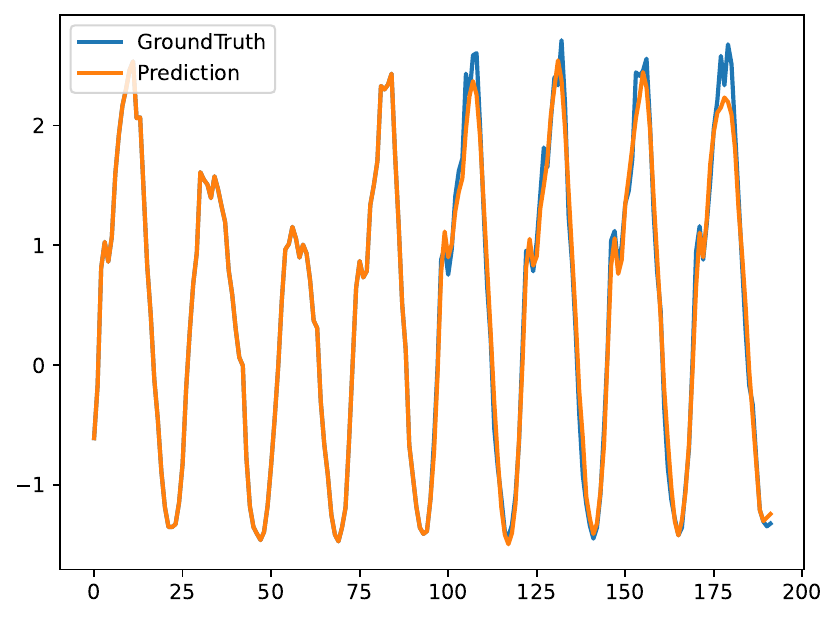}
		\caption*{Traffic}
	\end{minipage}
	\begin{minipage}[t]{0.26\textwidth}
		\centering
		\includegraphics[width=1.9in]{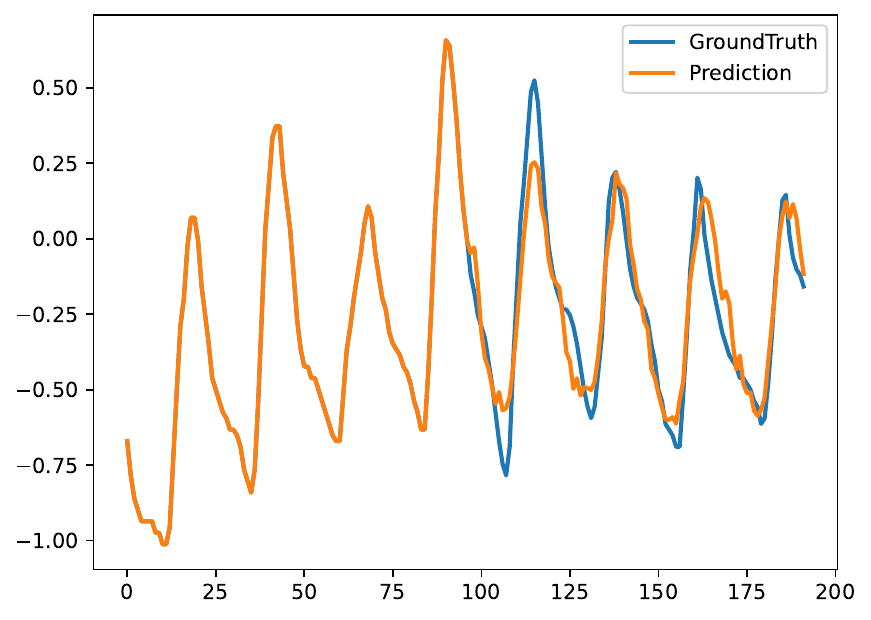}
		\caption*{ETTh2}
	\end{minipage}%
	\begin{minipage}[t]{0.26\textwidth}
		\centering
		\includegraphics[width=1.9in]{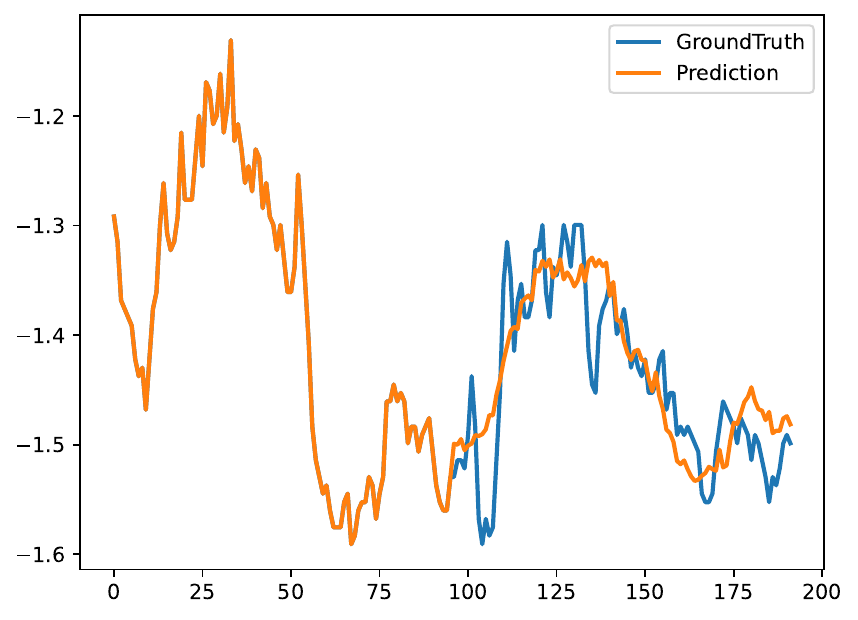}
		\caption*{ETTm1}
	\end{minipage}
	\begin{minipage}[t]{0.26\textwidth}
		\centering
		\includegraphics[width=1.9in]{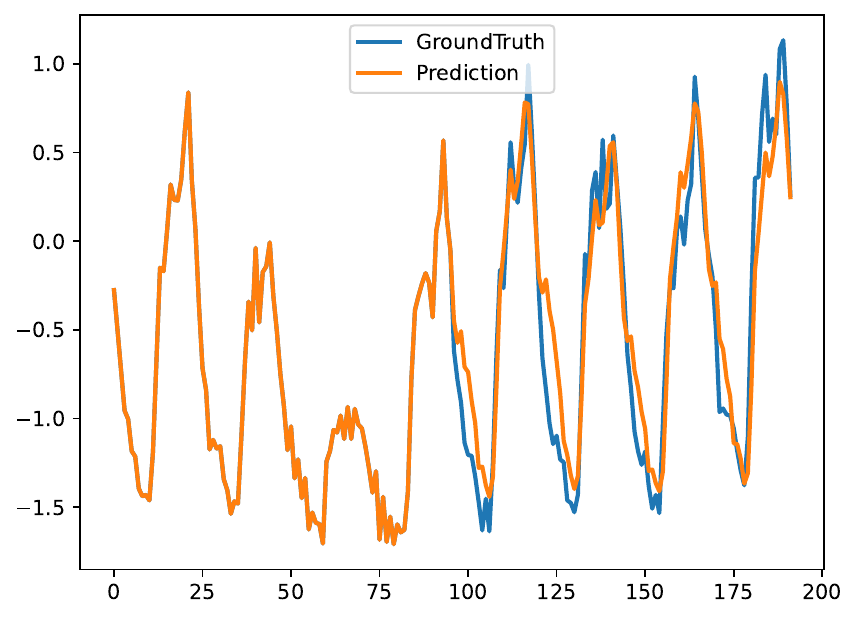}
		\caption*{Electricity}
	\end{minipage}%
	\begin{minipage}[t]{0.26\textwidth}
		\centering
		\includegraphics[width=1.8in]{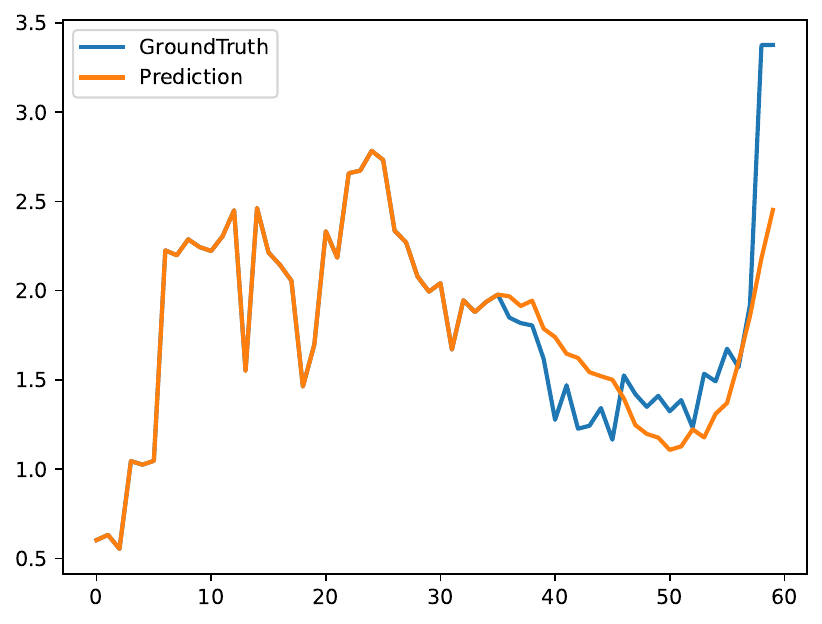}
		\caption*{ILL}
	\end{minipage}
\caption{Forecasting results of MPR-Net on multiple datasets exhibition}
\end{figure}

\section{Conclusion}

Based on the priori knowledge of pattern reproduction, this paper proposes a novel time series forecasting model, MPR-Net, which can be used for long-term and short-term forecasting tasks. The model is constructed as a hierarchical structure, and each layer contains three main blocks (HFE, PEF, and FFR) for handling different aspects of pattern information at various time scales. Compared with other advanced methods, MPR-Net makes full use of the temporal dependencies of time series, which not only makes its main time complexity to $O(n)$, but also makes the forecasting process interpretable. Finally, we conduct comparative experiments on more than 10 real data sets, and MPR-Net achieves state-of-the-art forecasting performance. At the same time, the overall superiority of various experimental results further verifies the effectiveness and robustness of the MPR-Net model, which can improve a new solution for the time series forecasting task. In the future, we will further mine the spatio-temporal relationship based on the existing work for the forecasting of spatio-temporal data.

\ifCLASSOPTIONcompsoc
  \section*{Acknowledgments}
\else
  \section*{Acknowledgment}
\fi

This work is supported by the National Natural Science Foundation of China (NSFC) Joint Fund with Zhejiang Integration of Informatization and Industrialization under Key Project (Grant No.U22A2033) and NSFC (Grant No.62072281).

\ifCLASSOPTIONcaptionsoff
  \newpage
\fi




\bibliographystyle{IEEEtran}
\bibliography{cite}
%

%
\begin{IEEEbiography}[{\includegraphics[width=1in,height=1.25in,clip,keepaspectratio]{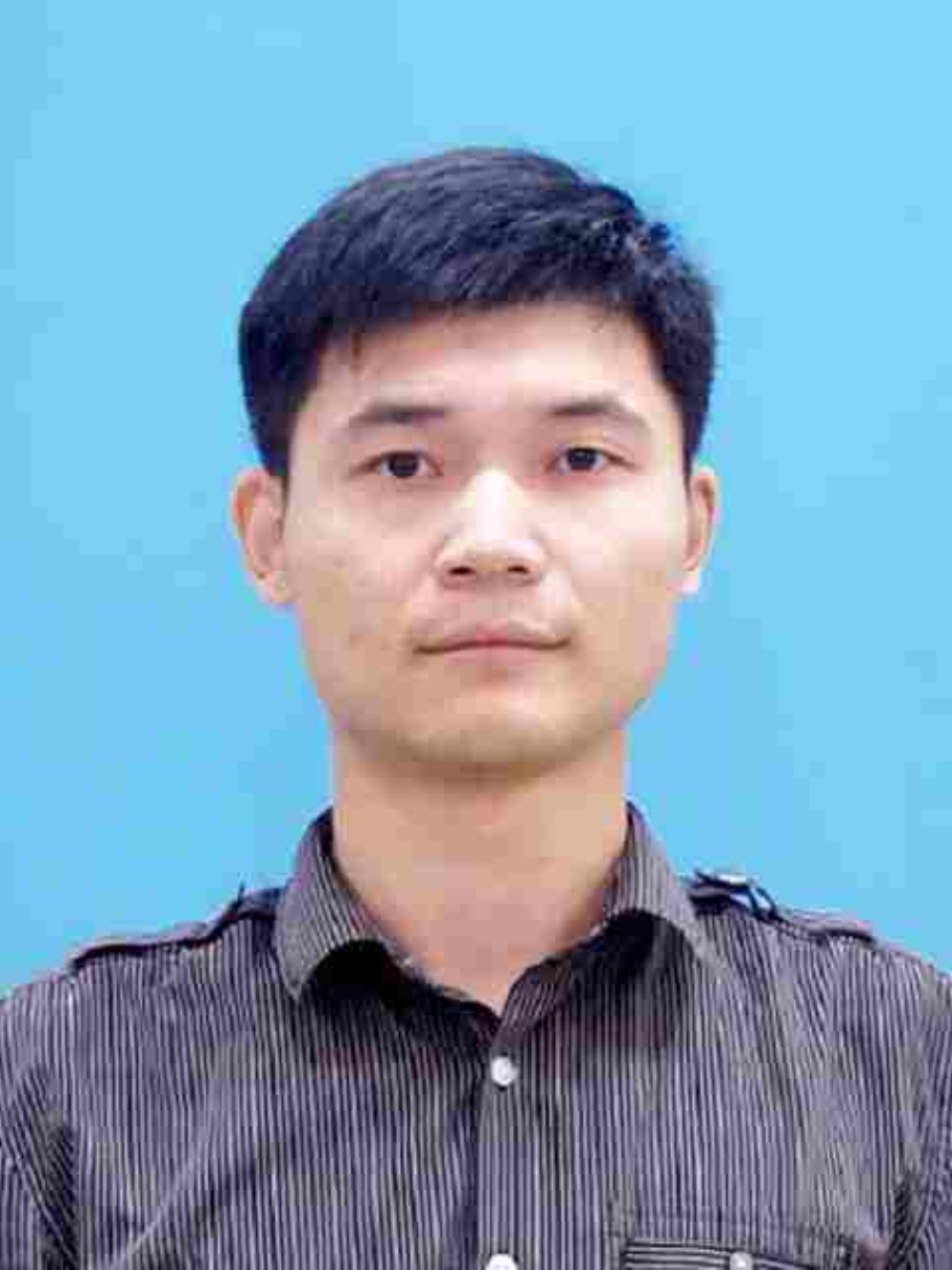}}]{Tianlong Zhao}
is currently a Ph.D. student in the School of Software, Shandong University, Jinan. He received the Master's degree from Shandong University of Finance and Economics, in 2019. His research interests include Reinforcement Learning, Deep Learning, Finance Intelligence and Data Mining.
\end{IEEEbiography}
\begin{IEEEbiography}[{\includegraphics[width=1in,height=1.25in,clip,keepaspectratio]{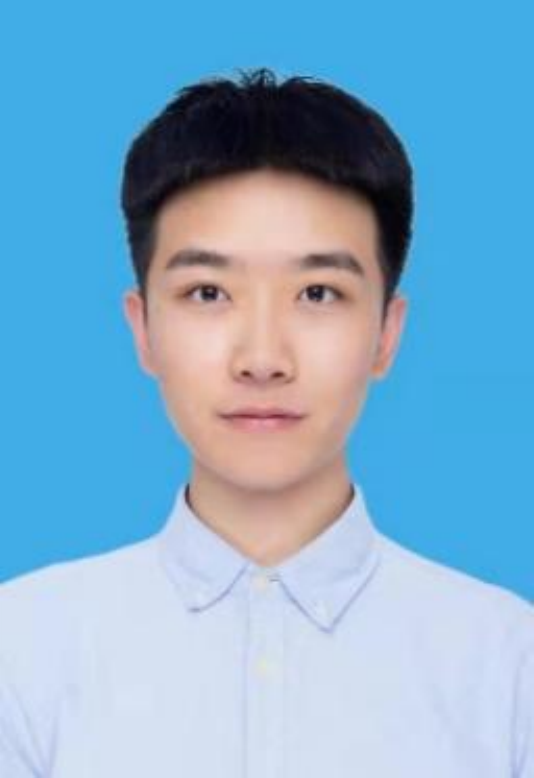}}]{Xiang Ma}
	is currently a Ph.D. student in the School of Software, Shandong University, Jinan. He received the Master's degree from Shandong University, in 2020. His research interests include image processing, finance intelligence and risk assessment.
\end{IEEEbiography}
\begin{IEEEbiography}[{\includegraphics[width=1in,height=1.25in,clip,keepaspectratio]{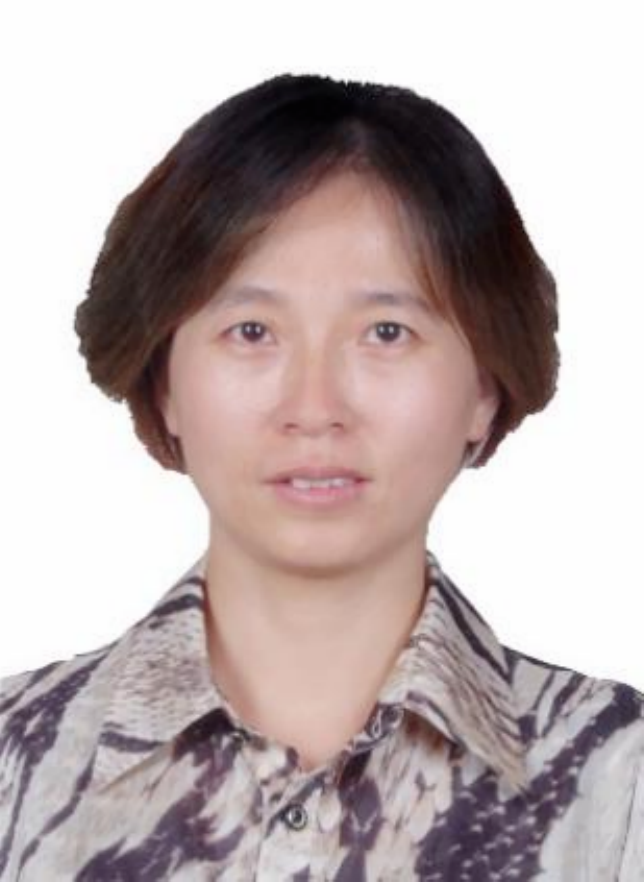}}]{Xuemei Li}
received the master’s and Ph.D. degrees from Shandong University, Jinan, China, in 2004 and 2010, respectively. From 2013 to 2014, she was a Visiting Scholar with the University of Houston, USA. She is currently a Professor with the School of Software, Shandong University, and a member of the intelligent computing and visualization Laboratory. She is engaged in research on CAGD, intelligent graphics/image processing, finance intelligence and risk assessment.
\end{IEEEbiography}
\begin{IEEEbiography}[{\includegraphics[width=1in,height=1.25in,clip,keepaspectratio]{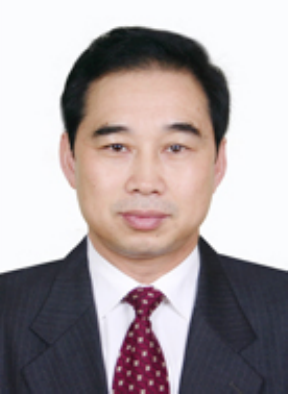}}]{Zhang Caiming}
	received the B.S. and M.S. degrees in computer science from Shandong University, Jinan, China, in 1982 and 1984, respectively, and the Ph.D. degree in computer science from the Tokyo Institute of Technology, Tokyo, Japan, in 1994. From 1998 to 1999, he was a Postdoctoral Fellow of the University of Kentucky, Lexington, USA. He is currently a Professor with the School of Software, Shandong University, and with the School of Computer Science and Technology, Shandong University of Finance and Economics. His research interests include CAGD, information visualization, finance intelligence and risk assessment.
\end{IEEEbiography}






\end{document}